\documentclass[10pt,twocolumn,letterpaper]{article}

\usepackage[pagenumbers]{cvpr} 

\definecolor{cvprblue}{rgb}{0.21,0.49,0.74}
\usepackage[pagebackref,breaklinks,colorlinks,allcolors=cvprblue]
{hyperref}
\usepackage{color}
\usepackage{textcomp}
\usepackage{stfloats}
\usepackage{url}
\usepackage{verbatim}
\usepackage{graphicx}
\usepackage{mathabx}
\usepackage{xcolor}
\usepackage{mathtools}
\usepackage{booktabs}
\usepackage{multirow}
\usepackage{tabu}
\usepackage{colortbl}
\usepackage{bm}

\newcommand{\blfootnote}[1]{%
  \begingroup
  \renewcommand\thefootnote{}\footnote{#1}%
  \addtocounter{footnote}{-1}%
  \endgroup
}

\def\paperID{31354} 
\def\confName{CVPR}
\def\confYear{2026}

\title{Scaling Dense Event-Stream Pretraining from Visual Foundation Models}

\author{Zhiwen Chen$^1$, Junhui Hou$^1$\textsuperscript{\dag}, Zhiyu Zhu$^1$,  Jinjian Wu$^2$\textsuperscript{\dag}, and Guangming Shi$^2$ \\
	City University of Hong Kong$^1$, Xidian University$^2$\\
	{\tt\footnotesize zhiwen.chen@cityu.edu.hk, jh.hou@cityu.edu.hk, jinjian.wu@mail.xidian.edu.cn}\\
    \url{https://github.com/zhiwen-xdu/ScaleEvent}}

\begin{document}
\hypersetup{urlcolor=magenta}

\twocolumn[{
    \vspace{-1.5cm}
    \maketitle
    \begin{center}
        \captionsetup{type=figure}
        \resizebox{1.0\textwidth}{!}{\includegraphics[trim={0 0 0 0},clip]{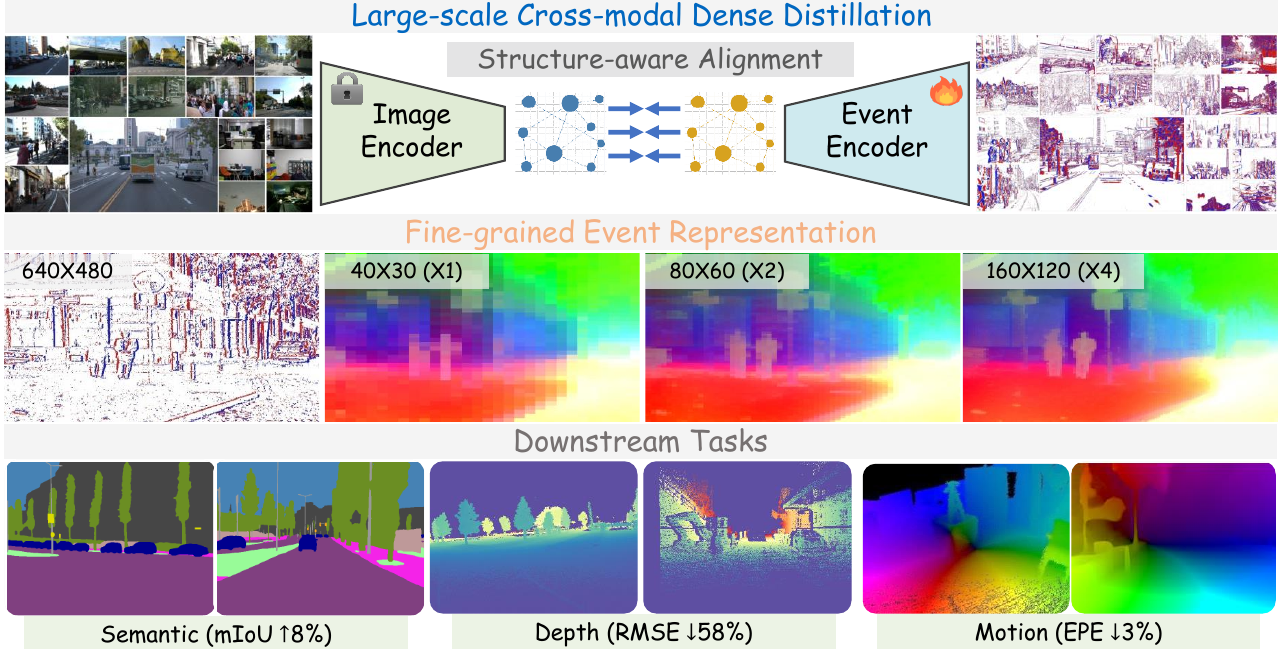}}
        \captionof{figure}{\textbf{ScaleEvent}: building upon large-scale cross-modal knowledge distillation from visual foundation models, we represent a novel pretraining method to scale up event representations. By anchoring dense cross-modal correspondences with a structure-aware loss, we obtain high-quality, fine-grained event representations that exhibit strong generalization across downstream dense perception tasks.}
        \label{fig:Cover}
        \vspace{-0.1cm}
        \label{fig:1}
    \end{center}
}]

\begin{abstract}
\blfootnote{Corresponding author$^{\dag}$: Junhui Hou and Jinjian Wu.}
Learning versatile, fine-grained representations from irregular event streams is pivotal yet nontrivial, primarily due to the heavy annotation that hinders scalability in dataset size, semantic richness, and application scope. To mitigate this dilemma, we launch a novel self-supervised pretraining method that distills visual foundation models (VFMs) to push the boundaries of event representation at scale. Specifically, we curate an extensive synchronized image-event collection to amplify cross-modal alignment. Nevertheless, due to inherent mismatches in sparsity and granularity between image-event domains, existing distillation paradigms are prone to semantic collapse in event representations, particularly at high resolutions. To bridge this gap, we propose to extend the alignment objective to semantic structures provided off-the-shelf by VFMs, indicating a broader receptive field and stronger supervision. The key ingredient of our method is a structure-aware distillation loss that grounds higher-quality image-event correspondences for alignment, optimizing dense event representations. Extensive experiments demonstrate that our approach takes a great leap in downstream benchmarks, significantly surpassing traditional methods and existing pretraining techniques. This breakthrough manifests in enhanced generalization, superior data efficiency and elevated transferability. 

\end{abstract}    
\vspace{-0.5cm}
\section{Introduction}
\label{sec:intro}
Event cameras, also referred to as bio-inspired vision sensors~\cite{lichtsteiner2006128,posch2010qvga,delbruck2010activity,gallego2020event}, fundamentally diverge from conventional frame-based cameras, distinguished by their attributes of ultra-low latency, high dynamic range, and minimal power consumption. Although the field of event-based scene understanding remains in its infancy, a rich spectrum of scene-specific applications has flourished~\cite{zhu2023cross,shiba2024secrets,wu2024motion,duan2025eventaid}.

As a prerequisite to comprehensive perception, learning high-quality and fine-grained event representations serves as the essential foundation. The prevalent paradigm relies on fully supervised training with dense event annotations. However, the irregular and labor-intensive nature of dense event labeling~\cite{jia2023event,xia2023cmda,li2024event,kong2025eventfly} severely constrains scalability. To mitigate this heavy annotation burden, several semi-~\cite{wang2021dual,wang2021evdistill,sun2022ess} and weakly-supervised~\cite{cho2024finding,jing2024hpl} approaches have been explored. Despite their promise, these methods remain constrained by limited pseudo-label quality and diversity, as well as ambiguous fine-grained features due to insufficient guidance, ultimately leading to suboptimal generalization in complex scenarios. Another compelling alternative is self-supervised learning, which leverages label-free pretext objectives to pretrain networks. By transferring established image-domain paradigms~\cite{he2022masked,he2020momentum,caron2021emerging,leeself}, event-based self-supervision learning has driven substantial progress. Nevertheless, intrinsic scarcity, discreteness, and sparsity of event data still impede model scaling and fine-grained representation quality. In this work, we mitigate these challenges through large-scale cross-modal knowledge distillation from visual foundation models.

Cross-modal knowledge distillation (KD) has recently emerged as a promising approach for unsupervised representation learning. In contrast to event-only self-supervision methods, which are constrained by intricate pretext task designs to exploit implicit dense patterns, KD furnishes stronger and richer proxy supervision, thereby substantially alleviating dependence on vast unlabeled datasets~\cite{liu2024small}. By leveraging shared knowledge from pretrained teachers, particularly powerful visual foundation models (VFMs), student models directly inherit strong semantic priors. Building on this insight, we propose a scalable event-based pretraining framework that distills VFMs to advance fine-grained event representations (as Fig.~\ref{fig:1}). Specifically, we construct an extensive collection of synchronized image-event pairs, spanning diverse conditions, including static versus ego-motion (motions), outdoor versus indoor (scenes), real-world versus simulation (sources), various event cameras (sensors), and multiple resolutions, aggregated from over ten large-scale datasets. This enables comprehensive cross-modal dense distillation, unlocking versatile and transferable event representations.

At its core, cross-modal knowledge distillation hinges on selecting high-quality image-event feature objectives for alignment. Existing methods can be categorized by the granularity at which their losses discriminate representations: pixel-/patch-level~\cite{chen2024segment}, or superpixel-/region-level~\cite{kong2024openess}. However, due to intrinsic discrepancies in sparsity and granularity between the image and event domains, these distillation losses indicate significant misalignment in representation spaces, leading to semantic collapse in the event domain, particularly at high resolutions. Specifically, pixel-level or patch-level alignment exacerbates mismatches, and superpixel-level methods depend on ambiguous fine-grained groups, amplifying erroneous guidance. To address these mismatches, we eschew meaningless event-image objectives and extend targets beyond brittle patch- and superpixel-level cues to discriminative semantic structures provided off-the-shelf by VFMs. This image-derived semantic structure expands the effective receptive field and delivers stronger and richer supervision, furnishing a comprehensive objective for event-image alignment. Concretely, we introduce a heuristic event-activation mask to regularize distillation toward informative regions, and we propose a structure-aware distillation loss that groups event-image correspondences over a broader field to suppress spurious matches. To achieve this, our alignment objective imposes structural constraints that steer event features toward image-consistent geometry. We optimize complementary intra- and cross-modal structure losses during pretraining, enabling more reliable representation learning.

Leveraging fine-grained event representations beyond the pretraining phase, our models transfer effectively to diverse dense perception tasks, including semantic segmentation, depth estimation, and optical flow estimation, consistently pushing task accuracy to new heights. Experimentally, this breakthrough significantly enhances performance, fostering improved generalization, superior data efficiency, and greater transferability.

In summary, the main contributions of this work are three-fold:
\begin{itemize}
\item we propose a novel self-supervised pretraining method that distills visual foundation models to scale up the boundaries of fine-grained event representations; 
\item we revisit event-domian semantic collapse in cross-modal distillation arising from image–event mismatches, and introduce a structure-aware alignment loss to regularize the pretraining process, thereby facilitating more reliable representation learning;
\item we demonstrate state-of-the-art performance across all evaluation settings and downstream dense perception tasks, with consistent gains in generalization, data efficiency, and transferability.
\end{itemize}

\section{Related Work}
\label{sec:related}


Representation learning underpins visual understanding. Driven by evolving perceptual demands, pretrained models have advanced in scale, versatility, and generalization. From a pretraining perspective, approaches fall into three regimes: fully supervised learning on large-scale data~\cite{dosovitskiy2020image,liu2021swin}, weakly supervised learning with reduced annotation requirements~\cite{singh2022revisiting,dehghani2023scaling,gao2024enhancing}, and self-supervised learning that harnesses intrinsic data features without relying on labels~\cite{he2020momentum, caron2021emerging, he2022masked}. Given the irregular and labor-intensive challenges of fine-grained event annotation, we focus our review on recent self-supervised methods across both image and event domains.

\subsection{Self-Supervised Visual Pre-training}

\noindent \textbf{Image Self-Supervision}. Self-supervised learning (SSL) has emerged as a powerful paradigm for visual pretraining. By learning directly from raw pixels and exploiting natural co-occurring patterns in images, SSL enables large-scale training. Learning without annotations requires auxiliary pretext tasks for surrogate supervision, with the core of SSL being the design of such tasks. \textbf{Due to the continuous nature of images}, early attempts derived supervisory from within the image, such as predicting relative patch position, re-ordering patches, re-colorizing, estimating transformations, or inpainting. Among these, inpainting-based methods gained traction with patch-based vision transformers~\cite{dosovitskiy2020image}, aiming to reconstruct corrupted regions as denoising autoencoders~\cite{he2022masked}. Subsequent works~\cite{baevski2023efficient, assran2023self} extended this idea to latent space, yielding richer representations. Meanwhile, another direction leveraged discriminative signals across images or patches, with advancements in contrastive learning~\cite{he2020momentum}, information-theoretic criteria~\cite{grill2020bootstrap}, self-distillation~\cite{caron2021emerging} and self-clustering~\cite{caron2020unsupervised}, all demonstrating strong feature learning. More recently, DINOv3~\cite{simeoni2025dinov3} pushed fine-grained representations and model scale further through extensive data collection and refined regularization.

\vskip 0.05in

\noindent \textbf{Event Self-Supervision.} By adapting 	established image pretraining paradigms, such as masked modeling~\cite{huang2024data,klenk2024masked}, contrastive learning~\cite{yang2023event}, and self-distillation~\cite{yang2024event}, event-based approaches have substantially advanced the field. Specifically, DMM~\cite{huang2024data} and MEM~\cite{klenk2024masked} employed masked modeling to reconstruct missing event parts. Likewise, ECDDP~\cite{yang2024event} grouped patch features into discriminative contexts and enforced context alignment. EUDA~\cite{ jian2023unsupervised} applied contrastive learning to cluster intra-object features while separating inter-object ones. In parallel, RLI~\cite{zhu2025revealing} drew inspiration from image denoising to reveal event latents. Moreover, STP~\cite{liang2025efficient} improved pretraining efficiency by fusing local-global contexts through prompt strategies. And TESPEC~\cite{mohammadi2025tespec} extended temporal modeling techniques from video pretraining to event cameras by exploiting long-term event sequences. Taken together, these advancements have markedly accelerated event representation learning.

\vskip 0.05in

\noindent \textbf{Remark.} Lessons from image-based self-supervision help event-based methods sidestep many pitfalls. Nevertheless, their fine-grained representation capability remains limited by two unresolved bottlenecks. First, insufficient data scale hinders knowledge emergence. Second, the discrete, sparse nature of event data complicates the design of pretext tasks that reliably exploit intrinsic dense patterns. In this work, we mitigate these challenges via cross-modal knowledge distillation from visual foundation models.

\subsection{Cross-modal Knowledge Distillation} 
Knowledge distillation (KD) transfers supervisory signal from a teacher to a student by training the latter to mimic the former’s behavior, such as outputs or intermediate features. Initially applied to compress large networks into smaller ones~\cite{hinton2015distilling,liu2024small}, KD has recently been revisited for semi-supervised~\cite{bartolomei2025depth,yang2025knowledge} and unsupervised~\cite{sautier2022image,liu2023segment,zhang2024fine,wu2025cm3ae,liu2025i2ekd} representation learning.

\vskip 0.05in

\noindent \textbf{Image-to-Event Knowledge Distillation.} Our work relates to KD from a pre-trained image teacher into an event student network. Early methods, such as E2VID~\cite{rebecq2019high}, extracted event representations by reconstructing events into grayscale images. This insight inspired later works like Evdistill~\cite{wang2021evdistill}, which bridges unpaired image-event data via bidirectional reconstruction, and ESS~\cite{sun2022ess}, which aligns event features to reconstructions to capture pixel-level detail. In parallel, ECDP~\cite{yang2023event} employed an image-event contrastive objective to learn scene-level context. With the advent of pre-trained vision-language models (e.g., CLIP~\cite{radford2021learning}), subsequent works~\cite{cho2023label,wu2023eventclip,zhou2024eventbind,xu2024ceia,li2025semantic,yang2025ezsr,li2025eventvl} advanced open-world event understanding through cross-modal distillation, though remaining confined to scene-level perception. More recently, fine-grained image-event distillation has gained traction. For instance, EventSAM~\cite{chen2024segment} distilled SAM~\cite{kirillov2023segment} to acquire semantic-agnostic, patch-level representations. DepthAnyEvent~\cite{bartolomei2025depth} and EventDAM~\cite{zhu2025depth} aligned proxy depth from DAv2~\cite{yang2024depth} to derive stereo-aware features. And OpenESS~\cite{kong2024openess} grouded superpixel-level multi-modal features by combining expert models (SAM + CLIP). Despite substantial progress, several limitations remain, namely, constrained distillation scale, task-specific objectives that impede scalability, and feature degradation arising from event heterogeneity, all of which we address through a unified dense pretraining framework.

\vskip 0.05in

\noindent \textbf{Remark.} Event-only self-supervision demands vast data with auxiliary pretext tasks to uncover intrinsic patterns. In contrast, cross-modal KD enables the student to inherit strong priors and richer proxy supervision from a pretrained teacher, substantially reducing reliance on large unlabeled datasets~\cite{liu2024small}. Moreover, the required multi-modal data are readily available from abundant unlabeled cross-modal collections or synthesized via VID2E simulation~\cite{gehrig2020video,hu2021v2e}, making cross-modal KD a feasible and scalable paradigm for event representation pretraining.

\section{Methodology}
\label{sec:method}
Our goal is to learn expressive fine-grained event representations through self-supervised distillation from vision foundation models such as DINOv3~\cite{simeoni2025dinov3}, leveraging the availability of aligned image and event data, without annotation requirements.

\begin{figure}
    \centering
    \includegraphics[clip, width=0.48\textwidth]{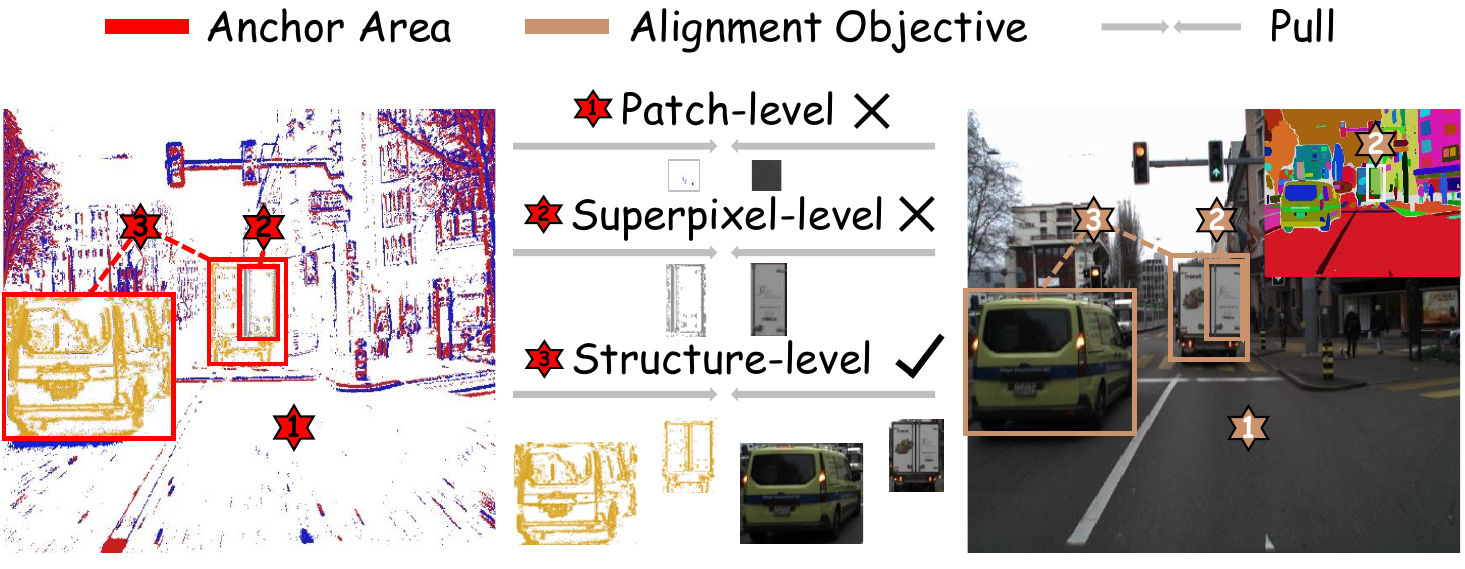}
    \caption{\textbf{Illustration of event-image feature alignment across granularities.} Patch-level alignment exacerbates cross-modal mismatches, superpixel grouping is ambiguous, while semantic structure grounds superior event–image correspondences.}
    \label{fig:Alignment}
    \vspace{-0.3cm}
\end{figure}

\subsection{Preliminaries}
\noindent \textbf{Synchronized Event-image Data.} Let $\mathcal{E}=\left\{(x,y,p,t)\right\} \in \mathbb{R}^{N \times 4 }$ represent a raw event set captured in a scene by an event camera during the time interval $t\rightarrow t+\Delta t$. The spatially and temporally synchronized image, denoted as $\boldsymbol{I} \in \mathbb{R}^{H \times W \times 3 }$, is captured by an RGB camera in the same scene at time $t$. Notably, the varying distribution of events challenges the stable sampling of edge-preserving event sets that align well with images. To enhance event input, we employ the motion-adaptive sampling algorithm in CrossEI~\cite{chen2024crossei}. Furthermore, to make events compatible with the vision foundation models, we aggregate the event set $\mathcal{E}$ into a three-dimensional volume $\boldsymbol{E} \in \mathbb{R}^{ H \times W \times B}$, following the setting in~\cite{zhu2019unsupervised,chen2024segment}. In our experiment, we set $B = 3$.

\vskip 0.05in

\noindent \textbf{Cross-modal Distilling.} To this end, we exploit the aligned and synchronized event and image data. Let $\mathbf{K}_n =\mathbf{F}_{\theta_e}(\boldsymbol{E}_n): \mathbb{R}^{ H \times W \times B} \mapsto \mathbb{R}^{H' \times W' \times D}$ be an event-based feature encoder with trainable parameters $\theta_e$, which takes as input an event volume $\boldsymbol{E}_n$ and outputs $D$-dimensional tokens $\mathbf{K}_n$ of downsampled spatial sizes $H'$ and $W'$. Our goal is to pretrain this event encoder \textit{without} accessing any annotations. Meanwhile, we integrate pre-trained DINOv3’s image encoder $\mathbf{Q}_n = \mathbf{G}_{\theta_i}(\boldsymbol{I}_n): \mathbb{R}^{ H \times W \times 3} \mapsto \mathbb{R}^{H' \times W' \times D}$ into distillation framework and keep the parameters $\theta_i$ fixed. In this context, we train $\mathbf{F}_{\theta_e}(\cdot)$ by aligning the event features $\mathbf{K}$ with the pre-trained image features $\mathbf{Q}$. In this work, we adopt a simple L1 loss for distillation:
\begin{equation}
\mathcal{L}_{\mathrm{l}_1}(\mathbf{K}, \mathbf{Q})
= \frac{1}{N}
\sum_{n=1}^{N}
\| \mathbf{K}_n - \mathbf{Q}_n \|_1,
\end{equation}
where $N$ is the number of cross-modal sample pairs in a mini-batch, $\mathbf{K}_n$, and $\mathbf{Q}_n$ are sample-wise tokens. 

\begin{figure}
    \centering
    \includegraphics[clip, width=0.471\textwidth]{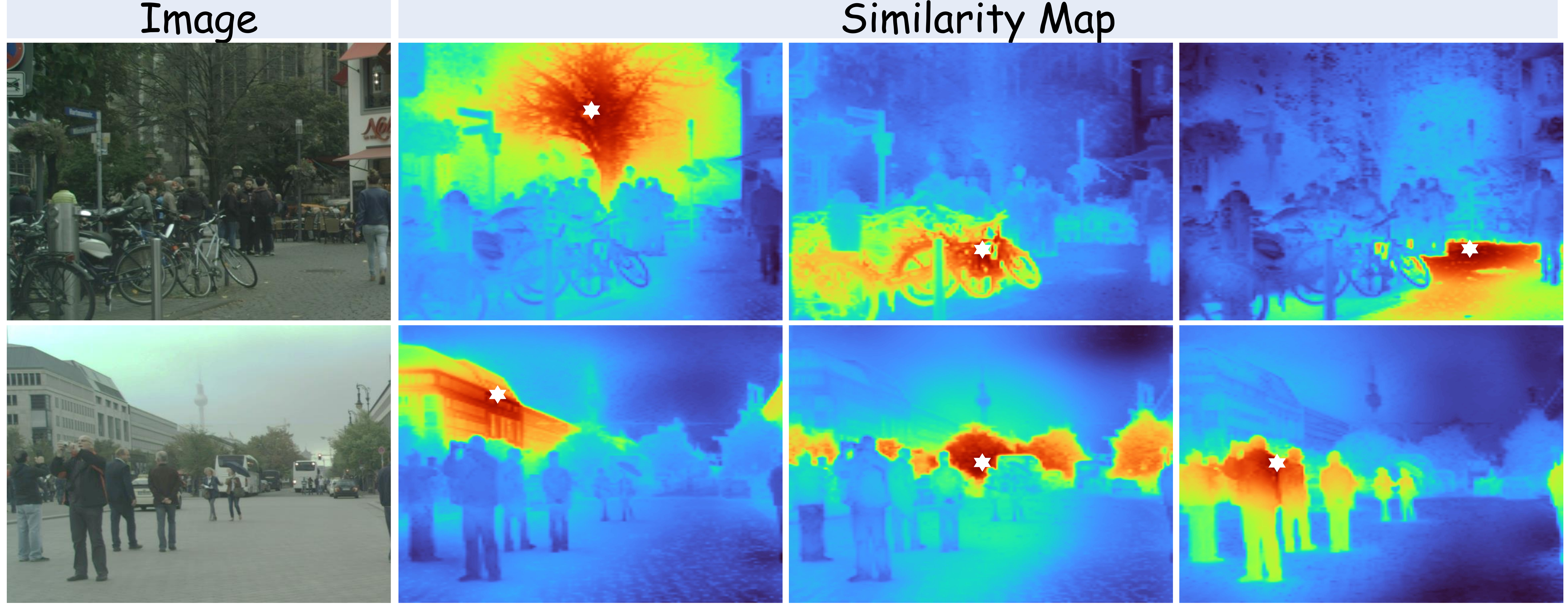}
    \caption{\textbf{Cosine similarity maps obtained with DINOv3 output features} (anchored at the distinct white stars). The image features exhibit coherent grouping induced by a strong off-the-shelf semantic structure.}
    \label{fig:Semantic}
    \vspace{-0.3cm}
\end{figure}

\begin{figure*}
\centering
    \resizebox{1.0\textwidth}{!}{\includegraphics[trim={0 0 0 0},clip]{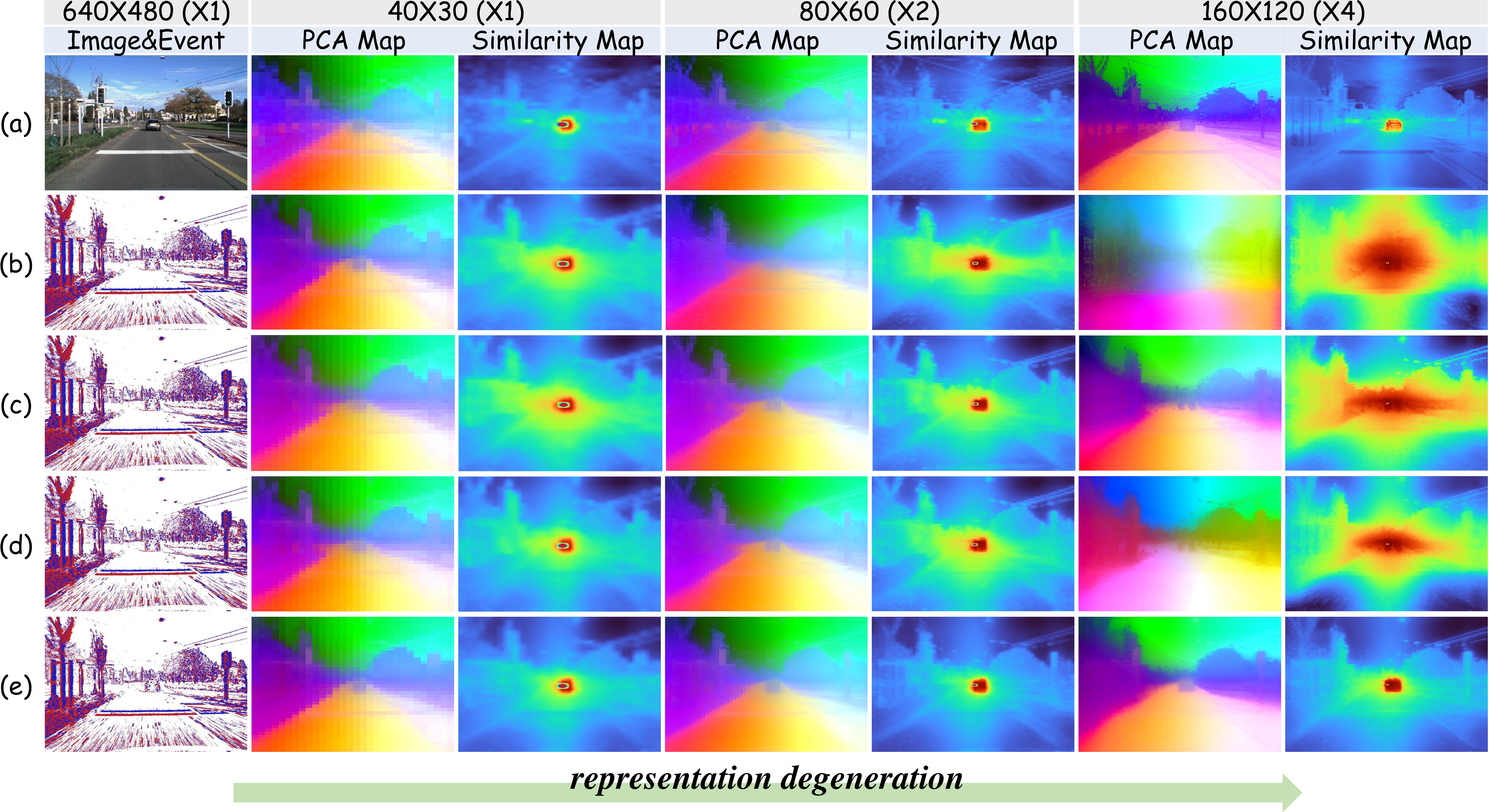}}
    \vspace{-0.3cm}
    \caption{\textbf{Comparison of dense event features under different distillation strategies}. All features are produced by a DINOv3-ViT-B model. \textbf{Left to right}: as spatial resolution increases, event representations degrade to varying degrees. PCA maps become less localized, and similarity maps (anchored at the \textcolor{red}{red dot}) become noisier. \textbf{Top to bottom}: (a) image features; (b) patch-level distillation; (c) superpixel-level distillation; (d) patch-level distillation + event activation mask; (e) patch-level distillation + event activation mask + structure-aware regularization (\textbf{Our method}). \emph{A more detailed feature analysis is provided in the supplementary materials.}}
    \label{fig:FeatureVis}
    \vspace{-0.4cm}
\end{figure*}

\subsection{Structure-aware Distillation Loss} 
\noindent \textbf{Event-domain Semantic Collapse.} Fundamentally, cross-modal knowledge distillation depends on well-posed image-event alignment objectives. As discussed in Sec.~\ref{sec:related}, fine-grained schemes are typically organized by loss granularity: patch-level or superpixel-level. However, the sparsity of event data versus the dense, texture-rich nature of images creates significant mismatches that render rigid correspondence losses prone to over-coupling (as illustrated in Fig.~\ref{fig:Alignment}). As revealed in Fig.~\ref{fig:FeatureVis}, this over-coupling distorts the representational geometry—suppressing local discriminability and precipitates semantic collapse of fine-grained event representations, an effect that intensifies with increasing resolution. To remedy these mismatches, we eschew meaningless image-event alignment objectives and extend the targets beyond patch- and superpixel-level cues to more discriminative semantic structures. Accordingly, we introduce an event-based activation mask that regularizes the distillation loss to favor informative image-event pairings, and we design a structure-aware distillation loss that groups feature correspondences across a broader receptive field, suppressing spurious matches.

\vskip 0.05in

\noindent \textbf{Activation Mask Constraint.} The sparsity of event data undermines fine-grained objectives: many event volume patches contain few or no events, yielding misleading alignment. We therefore concentrate on distilling high-activation event regions, where the signal is concentrated and motion texture is clearer, thereby improving supervision fidelity. Concretely, for each sampled event volume $\boldsymbol{E} \in \mathbb{R}^{ H \times W \times B}$, we compute an event density map $\boldsymbol{D} \in \mathbb{R}^{ H' \times W'}$ by summing patch-level events along the temporal axis: $\boldsymbol{D}(\mu, \nu) \;=\; \sum_{b=1}^{B}\sum_{(i,j) \in \mathcal{P}(\mu, \nu)} \,\phi\!\big(\boldsymbol{E}(i,j,b)\big)$,
where \( \mathcal{P} \) denotes the pixel indices that correspond to the patch at position \( (\mu, \nu) \), and $\phi(\cdot)$ maps activations to nonnegative counts (e.g., absolute value). We then derive a binary mask \(\mathbf{M}\in\{0,1\}^{H'\times W'}\) via applying a threshold $\tau$ to the density map:
\begin{equation}
\mathbf{M}(\mu, \nu)= \begin{cases}\mathbf{1}, & \text { if } \boldsymbol{D}(\mu, \nu) \geq \tau \\ \mathbf{0}, & \text { otherwise }\end{cases}
\end{equation}
with $\tau = 64$ controlling the high-activation area retained. This mask focuses the distillation objective on informative image-event feature pairs while suppressing spurious alignment in empty or low-activity regions.

\vskip 0.05in

\noindent \textbf{Structure-aware Alignment Loss.} Event data are dominated by dynamic edges. Although the interiors of object regions are sparse and information-poor, a larger receptive field reveals that these edge fragments coalesce into semantically coherent wholes. Treating these coherent structures as distillation objectives provides a principled bridge between discrete, sparse event data and dense, texture-rich images. As depicted in Fig.~\ref{fig:Semantic}, vision foundation models furnish an off-the-shelf prior, hereafter, \textbf{semantic structure}, that encodes similarity relations among all features (e.g., a pairwise affinity over tokens/features). This structure captures both local affinities and global dependencies, effectively enlarging the receptive field and delivering stronger and more stable supervision. Leveraging image-derived semantic structure thus supplies a comprehensive objective for aligning event and image representations. 

\vskip -0.01in

Accordingly, we distill this image-derived semantic structure into the event domain to ameliorate the inherent mismatch between the two modalities. We propose a regularization term that enforces structural consistency between the event and image representation spaces by penalizing discrepancies between their similarity graphs. The similarity graph is a weighted, undirected graph whose nodes are feature anchors (event or image tokens) and whose edges encode intra-modal pairwise affinities. In parallel, we use the event activation mask to select the anchored event feature, which suppresses background noise and strengthens the common semantic structure used for cross-modal alignment. Let $\mathbf{K}_n$ and $\mathbf{Q}_n$ denote event and image features of sample $n$ within the same batch, and $\mathbf{M}_n$ the corresponding event activation mask, with masked features denoted as $^*$. Using the shorthand $\mathbf{K}_n^* = \mathbf{K}_n \odot\mathbf{M}_n$ and $\mathbf{Q}_n^* = \mathbf{Q}_n \odot \mathbf{M}_n$, the intra-modal structure loss is defined as
\begin{small}
\begin{equation}
\mathcal{L}_{\text{is}}(\mathbf{K}^*,\mathbf{Q}^*) =\frac{1}{N} \sum_{n=1}^N \|(\mathbf{K}_n^*)(\mathbf{K}_n^*)^T-(\mathbf{Q}_n^*)(\mathbf{Q}_n^*)^T\|_1,
\end{equation}
\end{small}
To further reinforce structural consistency, we also penalize discrepancies between interactive similarity graphs that contrast predicted event-to-image affinities with source image-to-image affinities. This formulation compels each event feature’s similarity profile over all image features to mirror that of its paired image anchor, thereby aligning the event–image geometry with the image-domain structure. The cross-modal structure loss is as
\begin{small}
\begin{equation}
\mathcal{L}_{\text{cs}}(\mathbf{K}^*,\mathbf{Q}^*) =\frac{1}{N} \sum_{n=1}^N \|(\mathbf{K}_n^*)(\mathbf{Q}_n^*)^T-(\mathbf{Q}_n^*)(\mathbf{Q}_n^*)^T\|_1,
\end{equation}
\end{small}

Combining the masked L1 distillation term with the cross-modal and interactive structure-aware 
losses, we optimize the event encoder under the following objective:
\begin{small}
\begin{equation}
\mathcal{L}_{dis}=\mathcal{L}_{\mathrm{l}_1}(\mathbf{K}^*, \mathbf{Q}^*)+\lambda_{\text {is}} \mathcal{L}_{\text{is}}(\mathbf{K}^*,\mathbf{Q}^*)+\lambda_{\text {cs}} \mathcal{L}_{\text{cs}}(\mathbf{K}^*,\mathbf{Q}^*),
\end{equation}
\end{small}where $\lambda_{\text {is}} = 10$ and $\lambda_{\text {cs}} = 4$ are the regularization factors.
\section{Experiments}
\label{sec:exper}

\subsection{Pretraining Setup}

\noindent \textbf{Pretraining Datasets.} To pretrain a versatile and reliable event-based feature encoder $\mathbf{F}_{\theta_e}$ and scale up its parameters, we construct an extensive collection of synchronized image-event datasets, categorized by their source. \textbf{\emph{Real-world}}: DDD17~\cite{li2019event}, MVSEC~\cite{zhu2018multivehicle}, DSEC~\cite{gehrig2021dsec}, M3ED~\cite{Chaney_2023_CVPR}, VisEvent~\cite{wang2023visevent}, CoeSot~\cite{tang2025revisiting}, FEVD~\cite{kim2024frequency}, HighREV~\cite{sun2023event}, SEE-600K~\cite{lu2025see}. \textbf{\emph{VID2E synthetic}}~\cite{gehrig2020video}: GoPro~\cite{nah2019ntire}, SDSD~\cite{wang2021seeing}, DECD~\cite{rebecq2019high}, KITTI~\cite{geiger2013vision}, Cityscapes~\cite{cordts2016cityscapes}, Waymo~\cite{sun2020scalability}, DAVIS 2017~\cite{Pont-Tuset_arXiv_2017}. Due to the varying spatial resolutions of these datasets, we adopt the following processing strategy: setting the distillation resolution to $640 \times 480$, consolidating lower-resolution data and cropping higher-resolution data. After a series of data processing steps, we obtain a data collection of approximately $500K$ image-event pairs, all containing rich scene details. More dataset details are in the \textbf{\emph{supplementary materials}}.

\vskip 0.05in

\noindent \textbf{Feature Encoders.} We harness the state-of-the-art visual foundation model, DINOv3~\cite{simeoni2025dinov3}, as the teacher model for distillation. Our pretraings involve the ViT-S, ViT-B, and ViT-L versions, with a patch size of 16. For both the image and event processing branches, we employ identical feature encoders for knowledge distillation, initialized with DINOv3's pretrained weights.

\vskip 0.05in

\noindent \textbf{Implementation Details.} Our method is implemented using PyTorch. During pretraining, we fine-tune all parameters of the feature encoder. We utilize AdamW for optimization, setting the initial learning rate at $5 \times 10^{-6}$, with a momentum of $0.9$ and a weight decay of $1 \times 10^{-4}$.  The event encoder is pretrained for 10 epochs on four NVIDIA A6000 GPUs, with 100K event-image pairs per epoch. No data augmentation strategies are applied during pretraining.

\begin{table*}[htbp]
\captionsetup{skip=0pt}
\caption{Comparative study of different semantic segmentation methods under the linear probing (LP), few-shot fine-tuning, and full supervision (Full) settings, respectively, on the DDD17-Seg and DSEC-Semantic datasets. All mIoU scores are in percentage ($\%$). 
The \textbf{best mIoU} scores from each learning configuration are highlighted in \textbf{bold}.}
\centering
\renewcommand{\arraystretch}{1.0}
\setlength{\tabcolsep}{4.5pt}
\footnotesize
\begin{tabular}{c| c | c c c c c c | c c c c c c}
    \midrule[1.1pt]
    \multirow{2}{*}{\textbf{Method}} &\multirow{2}{*}{\textbf{Backbone}}  &\multicolumn{6}{c|}{\textbf{DDD17-Seg}} &\multicolumn{6}{c}{\textbf{DSEC-Semantic}}  \\
    \cline{3-14}
    \multicolumn{1}{c|}{} &  & LP & 1\% & 5\% & 10\% & 20\% & Full & LP & 1\% & 5\% & 10\% & 20\% & Full \\
    \midrule[1.1pt]
    MaskCLIP~\cite{zhou2022extract} &ViT-B/16 &31.91 & 53.91 & 56.27 & 59.32 & 59.97 & 61.27 & 33.08 & 33.89 & 37.03 & 38.83 & 42.40 & 55.01  \\
    FC-CLIP~\cite{yu2023convolutions} &ConvNeXt-L & 54.07 & 56.38 & 58.50 & 60.05 & 60.85 & 62.01 & 43.00 & 39.12 & 43.71 & 44.09 & 47.77 & 55.67  \\
    OpenESS~\cite{kong2024openess} &E2VID & 55.61 & \textbf{57.58} & 59.07 & 61.03 & 61.78 &\textbf{63.00} & 44.26 & 41.41 & 44.97 & 46.25 & 48.28 & 57.21  \\
    \midrule[1.1pt]
    \textbf{Ours} &ViT-B/16 &\textbf{57.87} &57.23 &\textbf{59.54} &\textbf{61.45} &\textbf{62.06} &62.81 &\textbf{58.42} & \textbf{54.37} &\textbf{62.82} &\textbf{63.88} &\textbf{64.15} & \textbf{64.93} \\
    \bottomrule[1.1pt]
\end{tabular}
\label{tab:2}
\vskip -0.05in
\end{table*}

\begin{table}[thbp]
\captionsetup{skip=0pt}
\caption{\textbf{Quantitative comparison of semantic segmentation} on the DDD17-Seg~\cite{alonso2019ev} and DSEC-Semantic dataset~\cite{sun2022ess} datasets. All scores are in percentage ($\%$). The best are marked with \textbf{bold}, and the second best are marked with \underline{underline}.}
\centering
\renewcommand{\arraystretch}{1.15}
\setlength{\tabcolsep}{1.1pt}
\footnotesize
\begin{tabular}{c| c | c |c c | c c}
    \midrule[1.1pt]
    \multirow{2}{*}{\textbf{Method}} &\multirow{2}{*}{\textbf{Present at}} &\multirow{2}{*}{\textbf{Backbone}}  &\multicolumn{2}{c|}{\textbf{DDD17}} &\multicolumn{2}{c}{\textbf{DSEC}}  \\
    \cline{4-7}
    \multicolumn{1}{c|}{} & &  &Acc $\uparrow$ &mIoU $\uparrow$ &Acc $\uparrow$ &mIoU $\uparrow$ \\
    \midrule[1.1pt]
    \rowcolor{gray!10} \multicolumn{7}{c}{\textbf{RGB initialization + Fully-Supervised}}   \\
    Ev-SegNet~\cite{alonso2019ev} &CVPRW'19 &Xception & 89.76 & 54.81 & 88.61 & 51.76    \\ 
    E2VID~\cite{rebecq2019high} & TPAMI'19 &ResNet-18 & 85.84 & 48.47 & 80.06 & 44.08    \\ 
    DTL~\cite{wang2021dual} &ICCV'21 &ResNet-50 & - & 58.80 & - & -    \\ 
    PVT-FPN~\cite{wang2021pyramid} & ICCV'21 &ResNet-34  &\underline{94.28} & 53.89 & - & - \\
    EVDistill~\cite{wang2021evdistill} & CVPR'21 &ResNet-34  &- & 58.02 & - & - \\
    MaskCLIP~\cite{zhou2022extract} & ECCV'22 &ViT-B/16  &90.50 &61.27 & 89.81 &55.01 \\
    ESS~\cite{sun2022ess} & ECCV'22 &E2VID  &88.43 &53.09 & 84.17 & 45.38 \\
    ESS-Sup~\cite{sun2022ess} & ECCV'22 &E2VID  & 91.08 & 61.37 & 89.37 & 53.29 \\
    HMNet~\cite{hamaguchi2023hierarchical}  &CVPR'23 &HMNet-L1  &- &- & 89.80 & 55.00 \\
    EvSegformer~\cite{jia2023event}  &TIP'23 &MiT-B1  &\textbf{94.72} & 54.41 &- &- \\
    FC-CLIP~\cite{yu2023convolutions} & NeurIPS'23 &CNeXt-L & 90.68 & 62.01 & 89.97 & 55.67 \\
    HALSIE~\cite{biswas2024halsie} & WACV'24 & SNN-ANN & 92.50 & 60.66 & 89.01 & 52.43 \\
    ESEG~\cite{zhao2025eseg} &AAAI'25 &MiT-B1 &90.68 &59.97 &\underline{91.47} & 57.55  \\
    KWYAF~\cite{li2025know} &AAAI'25 &MiT-B0 &91.32 &62.41 & 90.87 & 57.75  \\
    \rowcolor{blue!8} \multicolumn{7}{c}{\textbf{Event Pretraining + Fully-Supervised}}   \\
    ECDP~\cite{yang2023event} &ICCV'23 &ResNet-50 &- &59.15 &- &59.16  \\
    ECDDP~\cite{yang2024event} &ECCV'24 &ViT-S/16 &- &55.73 &- & 56.38  \\
    ECDDP~\cite{yang2024event} &ECCV'24 &Swin-T/7 &- &62.56 &- & 61.25  \\
    OpenESS~\cite{kong2024openess} &CVPR'24 &ResNet-50 &- &57.01 &- & 55.01  \\
    OpenESS~\cite{kong2024openess} &CVPR'24 &E2VID &91.05 & 63.00 & 90.21 & 57.21  \\
    STP~\cite{liang2025efficient} &ICCV'25 &ResNet-50 &- &62.13 &- &61.29  \\
    STP~\cite{liang2025efficient} &ICCV'25 &Swin-T/7 &- &\underline{63.29} &- &\underline{62.05}  \\
    
    \midrule[1.1pt]
    Ours  &- &ViT-S/16   &91.39   &59.64  &90.76  &61.12  \\
    Ours  &- &ViT-B/16   &92.21   &62.81  &92.00  &64.93  \\
    \textbf{Ours}  &- &ViT-L/16   &92.62   &\textbf{65.08}  &\textbf{93.10}  &\textbf{69.65}  \\
    \bottomrule[1.1pt]
\end{tabular}
\label{tab:1}
\vskip -0.05in
\end{table}

\subsection{Evaluation}
\subsubsection{Transfer Protocol}

\noindent \textbf{Task Decoders.} We evaluate the pretrained event feature encoder across diverse dense perception tasks, including semantic segmentation, monocular depth estimation, and optical flow estimation. By leveraging fine-grained event representations that closely align with image features, our encoder seamlessly integrates with image-domain decoders, enhancing scene perception. Specifically, we employ EoMT~\cite{kerssies2025your} for semantic decoding, DAv2~\cite{yang2024depth} for depth estimation, and SEA-RAFT~\cite{wang2024sea} for optical flow prediction. During downstream fine-tuning, all decoders are initialized from their released pretrained weights. 

\vskip 0.05in

\noindent \textbf{Benchmark Setup.} To probe data efficiency and transferability of our learned representations, we assess downstream performance. Beyond full supervision, we examine the linear probing (LP) and few-shot fine-tuning protocols tailored to tight parameter and annotation budgets. Under the linear probing setting, we optimize only the added task head and keep the weights of feature encoder frozen. In few-shot fine-tuning, we assume a very limited annotation budget, e.g., only $1\%$, $5\%$, $10\%$, or $20\%$ of the training set, with class-balanced, fixed-interval sampling and consistent optimization across methods. Together, these regimes assess data efficiency, robustness under limited resources, and cross-task transferability. More implementation details are in the \textbf{\emph{supplementary materials}}.

\begin{table*}[thbp]
\captionsetup{skip=0pt}
\caption{\textbf{Quantitative comparison of monocular depth estimation} on the MVSEC-Depth~\cite{zhu2018multivehicle} and DSEC-Depth~\cite{gehrig2021dsec} datasets. The best are marked with \textbf{bold}, and the second best are marked with \underline{underline}.}
\centering
\renewcommand{\arraystretch}{1.15}
\setlength{\tabcolsep}{2.0pt}
\footnotesize
\begin{tabular}{c | c| c | c c c c c c | c c c c c c}
    \midrule[1.1pt]
    \multirow{2}{*}{\textbf{Method}} &\multirow{2}{*}{\textbf{Present at}} &\multirow{2}{*}{\textbf{Backbone}}  &\multicolumn{6}{c|}{\textbf{MVSEC-Depth}} &\multicolumn{6}{c}{\textbf{DSEC-Depth}}  \\
    \cline{4-15}
    \multicolumn{1}{c|}{} & & &$\delta_1$ $\uparrow$ &$\delta_2\uparrow$\ &$\delta_3\uparrow$ &AbsRel$\downarrow$ &RMSE$\downarrow$ &RMSE log$\downarrow$ &$\delta_1\uparrow$ &$\delta_2\uparrow$\ &$\delta_3\uparrow$ &AbsRel$\downarrow$ &RMSE$\downarrow$ &RMSE log$\downarrow$ \\
    \midrule[1.1pt]
    \rowcolor{gray!10} \multicolumn{15}{c}{\textbf{RGB initialization + Fully-Supervised}}   \\
    E2Depth~\cite{hidalgo2020learning} &3DV'20 &ResNet-18 &0.432 &0.717 &0.868 &0.420 &7.268 &0.455 &0.409 &0.719 &0.891 &0.395 &13.258 &0.412  \\
    EReFormer~\cite{liu2024event}  &TVSVT'24 &Swin-T &0.391 &0.652 &0.810 &0.551 &8.373 &0.523 &0.524 &0.824 &0.945 &0.297 &11.608 &0.334  \\
    \rowcolor{blue!8} \multicolumn{15}{c}{\textbf{Event Pretraining + Fully-Supervised}}   \\
    ECDP~\cite{yang2023event}  &ICCV'23 &ViT-S/16 &0.476 &\underline{0.772} &0.863 &0.496 &7.680 &0.506 &0.528 &0.818 &0.938 &0.324 &11.473 &0.376  \\
    ECDDP~\cite{yang2024event}  &ECCV'24 &ViT-S/16 &\underline{0.513} &0.762 &0.871 &0.428 &6.957 &\underline{0.469} &0.545 &0.857 &0.959 &0.263 &9.477 &0.294  \\
    DepthAnyEvent-R~\cite{bartolomei2025depth} &ICCV'25 &ViT-S/16 &0.489 &0.751 &\underline{0.878} &\underline{0.365} &\underline{6.465} &0.483 &\underline{0.691} &\underline{0.930} &\underline{0.981} &\underline{0.191} &\underline{8.880} &\underline{0.266}  \\
    \midrule[1.1pt]
    Ours  &- &ViT-S/16 &0.577 &0.800 &0.914 &0.289 &6.145 &0.378 &0.824 &0.965 &0.993 &0.131 &4.564 &0.184 \\
    Ours  &- &ViT-B/16 &0.594 &0.811 &0.922 &0.280 &5.891 &0.364 &0.872 &0.978 &0.996 &0.109 &4.032 &0.158 \\
    \textbf{Ours} &- &ViT-L/16  &\textbf{0.625} &\textbf{0.834} &\textbf{0.934} &\textbf{0.268} &\textbf{5.554} &\textbf{0.343} &\textbf{0.896} &\textbf{0.983} &\textbf{0.997} &\textbf{0.101} &\textbf{3.694} &\textbf{0.144} \\
    \bottomrule[1.1pt]
\end{tabular}
\label{tab:3}
\end{table*}

\begin{table*}[thbp]
\captionsetup{skip=0pt}
\caption{Comparative study of different monocular depth estimation methods under the linear probing (LP) and few-shot fine-tuning, and full supervision (Full) settings, respectively, on the MVSEC-Depth and DSEC-Depth datasets. The \textbf{best RMSE} from each learning configuration are highlighted in \textbf{bold}.}
\centering
\renewcommand{\arraystretch}{1.0}
\setlength{\tabcolsep}{4.2pt}
\footnotesize
\begin{tabular}{c| c | c c c c c c | c c c c c c}
    \midrule[1.1pt]
    \multirow{2}{*}{\textbf{Method}} &\multirow{2}{*}{\textbf{Backbone}}  &\multicolumn{6}{c|}{\textbf{MVSEC-Depth}} &\multicolumn{6}{c}{\textbf{DSEC-Depth}}  \\
    \cline{3-14}
    \multicolumn{1}{c|}{} &  & LP & 1\% & 5\% & 10\% & 20\% & Full & LP & 1\% & 5\% & 10\% & 20\% & Full \\
    \midrule[1.1pt]
    DepthAnyEvent-R~\cite{bartolomei2025depth} &ViT-S/16 &7.473 & 7.542 &7.261 &6.794 &6.637 &6.465 & 10.584 & 10.347 &9.898 &9.534 & 9.065 & 8.880  \\
    \midrule[1.1pt]
    \textbf{Ours} &ViT-S/16 &\textbf{6.756} &\textbf{6.930} &\textbf{6.712} &\textbf{6.477} &\textbf{6.352} &\textbf{6.145} &\textbf{4.861} & \textbf{4.983} &\textbf{4.751} &\textbf{4.728} &\textbf{4.694} & \textbf{4.564} \\
    \bottomrule[1.1pt]
\end{tabular}
\label{tab:4}
\end{table*}

\subsubsection{Semantic Segmentation}

\noindent \textbf{Settings.} Following the setup of OpenESS~\cite{kong2024openess}, we evaluate the DDD17-Seg~\cite{alonso2019ev} and DSEC-Semantic~\cite{sun2022ess} datasets for semantic segmentation. Mean interaction over union (mIoU) and average class accuracy (Acc) are used as evaluation metrics. 

\vskip 0.05in

\noindent \textbf{Results.} We perform a thorough comparison of our pretrained model with RGB-based transfer methods and other event-domain pretraining approaches. As shown in Tab.~\ref{tab:1}, our method achieves the highest mIoU of $65.08\%$, and $69.65\%$ on the DDD17-Seg and DSEC-Semantic datasets, respectively, surpassing all event-domain segmentation models. Notably, on DSEC-Semantic, we outperform the recent SOTA model STP~\cite{liang2025efficient} by $7.6\%$, advancing the boundaries of event-based semantic representation.
In linear probing, our method reaches an mIoU of $58.42\%$, surpassing the best RGB-transfer method, KWYAF~\cite{li2025know} at $57.75\%$ (Tab.~\ref{tab:2}). For few-shot fine-tuning on DSEC-Semantic, we attain $62.82\%$ mIoU with just $5\%$ of the training data, outperforming OpenESS~\cite{kong2024openess} at $57.21\%$. This trend persists across other data proportions, with our method consistently leading or closely competing with the best results. Consistent gains over prior art attest to the efficacy and superiority of our pretraining strategy.

\subsubsection{Depth Estimation}

\noindent \textbf{Settings.} Following the setup of DepthAnyEvent~\cite{bartolomei2025depth}, we evaluate on the MVSEC-Depth~\cite{zhu2018multivehicle} and DSEC-Depth datasets~\cite{gehrig2021dsec} for monocular depth estimation. The depth error metrics are absolute relative error (AbsRel), root mean squared error (RMSE), logarithmic RMSE (RMSE log), and accuracy with different thresholds ($\delta<1.25$ ($\delta_1$), $\delta<1.25^2$ ($\delta_2$), and $\delta<1.25^3$ ($\delta_3$) ).

\vskip 0.05in

\noindent \textbf{Results.} As shown in Tab.~\ref{tab:3}, our method achieves the lowest depth estimation error, significantly outperforming current arts. Notably, on DSEC-Depth, we achieve $99.7\%$ $\delta_3$ accuracy and an RMSE of $3.694$. Using the same backbone, we reduce the RMSE of DepthAnyEvent-R~\cite{bartolomei2025depth} from $8.880$ to $4.564$, marking a substantial improvement. As shown in Tab.~\ref{tab:4}, the linear probing results highlight the pretrained model's robust event representation capability, with freezing the feature encoder having minimal impact on depth estimation. In the few-shot fine-tuning setting, we achieve an RMSE of $4.983\%$  with only $1\%$ of the annotation data. These consistent improvements over prior work underscore the efficacy of task-agnostic pretraining in capturing richer domain knowledge.

\subsubsection{Optical Flow Estimation}

\begin{table*}[thbp]
\captionsetup{skip=0pt}
\caption{\textbf{Ablative study results of the proposed key components}. \textbf{Distill} denotes the knowledge distillation mark; \textbf{Mask} refers to activation mask regularization; \textbf{IS Loss} denotes the intra-modal structure loss; \textbf{CS Loss} denotes the cross-modal structure loss; \textbf{(a)} indicates the use of a image-domain pretrained model; \textbf{(b)} serves as our baselines; \textbf{(f)} denotes our complete cross-modal distillation framework.}
\centering
\renewcommand{\arraystretch}{1.15}
\setlength{\tabcolsep}{5pt}
\footnotesize           
\begin{tabular}{c |c c c c | c c| c  c |c c | cc}
   \toprule
    \multirow{2}{*}{\textbf{Exp.}} &\multirow{2}{*}{\textbf{Distill}} &\multirow{2}{*}{\textbf{Mask}} &\multirow{2}{*}{\textbf{IS Loss}} &\multirow{2}{*}{\textbf{CS Loss}} &\multicolumn{2}{c|}{\textbf{DDD17-Seg}} &\multicolumn{2}{c|}{\textbf{DSEC-Semantic}} &\multicolumn{2}{c|}{\textbf{MVSEC-Depth}} &\multicolumn{2}{c}{\textbf{DSEC-Depth}}\\
    \cline{6-13}
    \multicolumn{1}{c|}{} & & & & &Acc$\uparrow$ &mIoU$\uparrow$ &Acc$\uparrow$ &mIoU$\uparrow$ &$\delta_1$ $\uparrow$  &RMSE $\downarrow$ &$\delta_1$ $\uparrow$  &RMSE $\downarrow$ \\
    \hline
    \textbf{(a)}  & & & &  &91.39 &59.60  &91.94 &64.31  &0.593  &6.635  &0.846 & 4.424    \\
    \textbf{(b)}  &$\checkmark$  &  & & &92.16  &62.41   &92.74 &66.17  &0.609  &6.114  &0.875 &4.063 \\
    \textbf{(c)}  &$\checkmark$ &$\checkmark$  & &  &92.32  &62.67   &92.82 &66.54  &0.611  &5.922  &0.876 &4.025 \\
    \textbf{(d)}  &$\checkmark$ &$\checkmark$ &$\checkmark$  & &92.60 &64.84  &93.08 &69.20  &0.620  &5.684  &0.889 &3.792   \\
    \textbf{(e)}  &$\checkmark$ &$\checkmark$ &  &$\checkmark$   &92.51 &63.62  &93.01 &68.68  &0.614  &5.786  &0.881 &3.870  \\
    \textbf{(f)}  &$\checkmark$ &$\checkmark$ &$\checkmark$  &$\checkmark$ &92.62  &65.08  &93.10  &69.65  &0.625  &5.554   &0.896  &3.694 \\
    \bottomrule
\end{tabular}
\label{tab:ablation}
\vskip -0.05in
\end{table*}

\begin{table}[h]
\captionsetup{skip=0pt}
\caption{\textbf{Quantitative comparison of optical flow estimation} on the MVSEC-Flow~\cite{zhu2018multivehicle} dataset. The best are marked with \textbf{bold}, and the second best are marked with \underline{underline}.}
\centering
\renewcommand{\arraystretch}{1.2}
\setlength{\tabcolsep}{1.0pt}
\footnotesize
\begin{tabular}{c| c | c c |c c | c c}
    \midrule[1.1pt]
    \multirow{2}{*}{\textbf{Method}} &\multirow{2}{*}{\textbf{Backbone}}  &\multicolumn{2}{c|}{\textbf{indoor flying1}} &\multicolumn{2}{c|}{\textbf{indoor flying2}} &\multicolumn{2}{c}{\textbf{indoor flying3}}  \\
    \cline{3-8}
    \multicolumn{1}{c|}{} &  &EPE $\downarrow$ &Out$\downarrow$ &EPE $\downarrow$ &Out$\downarrow$ &EPE $\downarrow$ &Out$\downarrow$ \\
    \midrule[1.1pt]
    \rowcolor{gray!10} \multicolumn{8}{c}{\textbf{RGB initialization + Fully-Supervised}}   \\
    EST~\cite{gehrig2019end} &ResNet-18 & 1.24 & 5.09 & 2.05 & 19.90 & 1.71 & 11.67    \\ 
    DCEFlow~\cite{wan2022learning} & - & 0.75 & 0.60 & 1.39 & 8.01 & 1.13 & 5.29    \\ 
    \rowcolor{blue!8} \multicolumn{8}{c}{\textbf{Event Pretraining + Fully-Supervised}}   \\
    ECDP~\cite{yang2023event} &ResNet-50 & 0.60 & 0.35 & 1.35 & 8.57 & 1.12 & 5.26  \\
    ECDP~\cite{yang2023event} & ViT-S/16 & 0.61 & 0.05 & 1.26 & 6.69 & 1.00 & 3.11  \\
    ECDDP~\cite{yang2024event} &Swin-T/7 &0.36 &0.04 &0.45 &0.002 &0.42 & 0.001 \\
    ECDDP~\cite{yang2024event} &ViT-S/16 &0.51 &0.11 &0.69 &0.29 &0.61 &0.08  \\
    STP~\cite{liang2025efficient} &Swin-T/7 &\underline{0.31} &\underline{0.03} &\underline{0.41} &\underline{0.001} & $\underline{0.43}$ &\underline{0.001} \\
    STP~\cite{liang2025efficient} &ViT-S/16 & 0.58 & 0.05 & 1.22 & 6.34 & 0.93 & 3.03 \\
    \midrule[1.1pt]
    Ours  &ViT-S/16   &\textbf{0.29} &\textbf{0.03} &\textbf{0.38} &\textbf{0.001} &\textbf{0.40} &\textbf{0.001} \\
    \bottomrule[1.1pt]
\end{tabular}
\label{tab:5}
\vskip -0.05in
\end{table}

\noindent \textbf{Settings.} Following the setup of ECDDP~\cite{yang2024event}, we evaluate event-based optical flow estimation on the MVSEC-Flow~\cite{zhu2018multivehicle} dataset. The metrics are the average endpoint error (EPE) and the outlier ratio ($\%$ Out), where pixels with an EPE above 3 and $5\%$ of the ground truth optical flow magnitudes are considered outliers. All measurements are taken over pixels with valid ground truth and at least one event.

\vskip 0.05in

\noindent \textbf{Results.} As shown in Tab.~\ref{tab:5}, our pretrained event encoder achieves the lowest average endpoint error and outlier ratio. Despite the ViT architecture not being inherently optimized for optical flow estimation, we still achieve performance comparable to SOTA. Additionally, we provide extensive optical flow evaluation results in the \textbf{\emph{supplementary materials}}, further demonstrating the strong generalization capability of our approach.

\subsubsection{Ablation Studies}

\noindent \textbf{Effect of Key Components.} We conduct a series of comprehensive experiments to uncover the interplay and effectiveness of the proposed loss items, as summarized in Tab.~\ref{tab:ablation}. All ablation experiments are based on our largest model, ViT-L, unless otherwise specified. Overall, the comparison between (a) and (f) demonstrates the significance of event pretraining, which significantly enhances the event-domain representation capability. Specifically, the contrast between (a) and (b) underscores that simple large-scale multi-modal alignment can substantially improve representation power, highlighting its effectiveness. After applying the activation mask constraint to the aligned regions ((b) versus (c)), all methods show consistent improvements. However, due to the lack of semantic awareness in the activation mask, this regularization is not fully optimal. The comparison between (c) and (d) (and (c) and (e)) reveals that aligning the feature space with semantic structure significantly enhances the pretrained model's performance, particularly with the intra-modal structure loss. Notably, the simultaneous application of both intra-modal and cross-modal structure losses results in significantly greater improvements than either method alone, demonstrating their complementary nature. While cross-modal structure loss alone yields only modest improvements, it still contributes to the overall superior performance of our pretrained models. 

For additional experimental results and qualitative assessments, please refer to the \textbf{\emph{supplementary materials}}.

\section{Conclusion}
In this study, we introduced a versatile self-supervised learning framework that enhances fine-grained event representations by promoting large-scale structure-aware image-event alignment. Additionally, our work revisits the use of visual foundation models (VFMs) to improve event-based scene understanding. Through extensive experimentation across a variety of downstream tasks, we demonstrated the effectiveness and superiority of our framework. We believe this research will serve as a catalyst for further integration of large-scale image and event representation learning, paving the way for the development of more robust, scalable, and annotation-efficient perception models. This approach holds significant promise for advancing the field of cross-modal perception and improving real-world applications in dynamic environments.

\clearpage
\setcounter{page}{1}
\maketitlesupplementary

In this appendix, we supplement the following materials to support the findings and observations in the main body of this paper:
\begin{itemize}
\item Section~\ref{appendix:implement} elaborates on detailed implementation specifics to facilitate reproduction;
\item Section~\ref{appendix:quantitative} presents the complete quantitative results of our experiments;
\item Section~\ref{appendix:qualitative} includes extensive qualitative results to indicate clearer visual comparisons;
\item Section~\ref{appendix:imitation} provides a further analysis of the current limitations and discusses potential improvement methods.
\end{itemize}

\section{Additional Implementation Detail}
\label{appendix:implement}

\subsection{Pretraining Datasets}
In this work, we assemble an extensive collection of synchronized image-event datasets to pretrain a versatile and reliable event-domain feature encoder. These datasets span diverse sensing conditions, motion patterns, environments, and acquisition pipelines, providing broad coverage for large-scale cross-modal alignment. A summary of the detailed configurations and salient characteristics of these pretrained datasets is shown in Table~\ref{tab:7} and Table~\ref{tab:8}, grouped by real-world and synthetic sources.

\begin{figure}
    \centering
    \includegraphics[clip, width=0.48\textwidth]{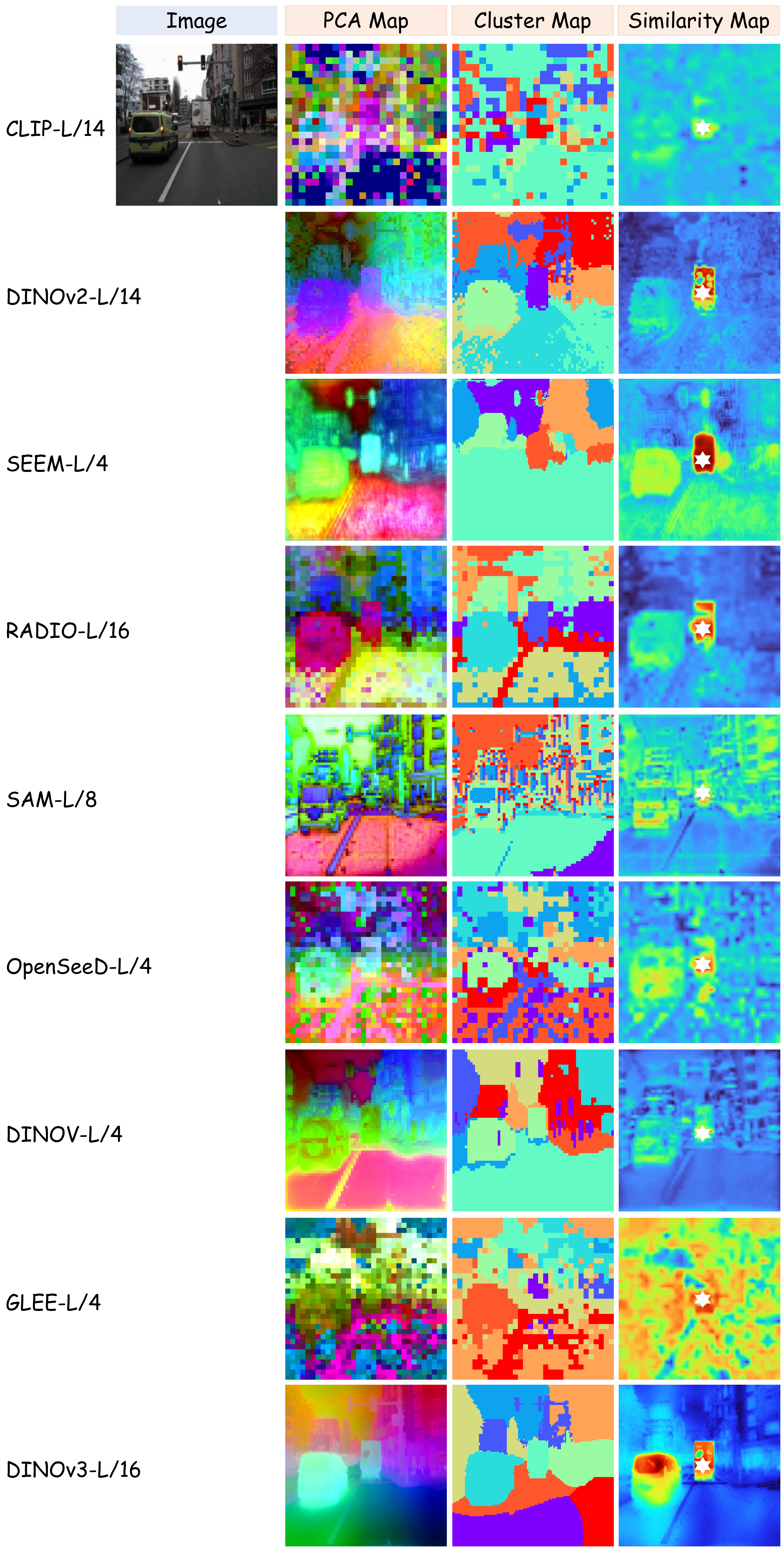}
    \vspace{-0.5cm}
    \caption{\textbf{Comparison of dense image features} under different visual foundation models through a toy example.}
    \label{fig:VFMs}
    \vspace{-0.5cm}
\end{figure}

\subsection{Vision Foundation Models}
In this work, we adopt the state-of-the-art visual foundation model DINOv3~\cite{simeoni2025dinov3} as the teacher model to distill fine-grained representations into our event encoder. Before committing to this choice, we conducted a brief comparative analysis of representative VFMs: CLIP~\cite{radford2021learning}, DINOv2~\cite{oquab2023dinov2}, SAM~\cite{kirillov2023segment}, SEEM~\cite{zou2023segment}, RADIO2.5~\cite{heinrich2025radiov2}, OpenSeeD~\cite{zhang2023simple}, DINOV~\cite{li2024visual}, GLEE~\cite{wu2024general}, and DINOv3, with emphasis on fine-grained representation fidelity (token-level affinities, boundary sharpness, and global-local coherence). Using a controlled toy example (Figure~\ref{fig:VFMs}), we probed the quality of the learned semantic structure. DINOv3 consistently exhibited the most coherent long-range grouping and the clearest region boundaries, and is therefore selected as our teacher model. Supporting qualitative results are reported in~\cite{simeoni2025dinov3}.

\begin{table*}
\caption{The pretraining dataset configuration and data statistics for the \textbf{nine real-world event-image datasets} used in our experiments.}
\centering
\renewcommand{\arraystretch}{1.15}
\setlength{\tabcolsep}{5.5pt}
\small
\begin{tabular}{l|l|l|l|l}
\hline Dataset & Illustration &Resolution & Statistics &Source\&Type \\
\hline DDD17~\cite{binas2017ddd17} &\raisebox{-.5\height}{\includegraphics[width=5.0cm,height=1.9cm]{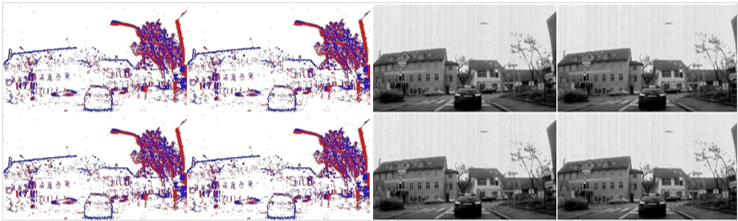}} & \begin{tabular}{l}
$346 \times 260$\\
\end{tabular} & \begin{tabular}{l}
5,000 pairs \\
$\approx$ 20 categories \\
36 sequences \\
\end{tabular} & \begin{tabular}{l}
Real-world \\
DAVIS346B \\
Low-resolution \\
Driving Scene \\
\href{https://pkuml.org/resources/pku-ddd17-car.html}{Download}
\end{tabular} \\
\hline MVSEC~\cite{zhu2018multivehicle} &\raisebox{-.5\height}{\includegraphics[width=5.0cm,height=1.9cm]{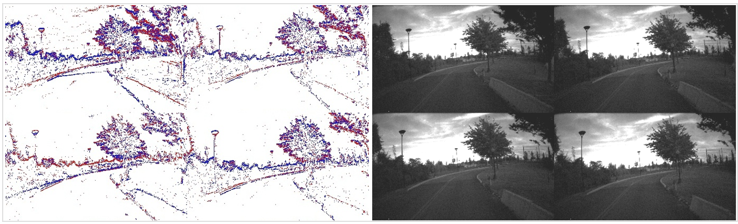}} & \begin{tabular}{l}
$346 \times 260$
\end{tabular} & \begin{tabular}{l}
5,000 pairs \\
$\approx$ 20 categories \\
9 sequences \\
\end{tabular} & \begin{tabular}{l}
Real-world \\
DAVIS346B \\
Low-resolution \\
Driving Scene \\
\href{https://daniilidis-group.github.io/mvsec/}{Download}
\end{tabular} \\

\hline SEE-600K~\cite{lu2025see} &\raisebox{-.5\height}{\includegraphics[width=5.0cm,height=1.9cm]{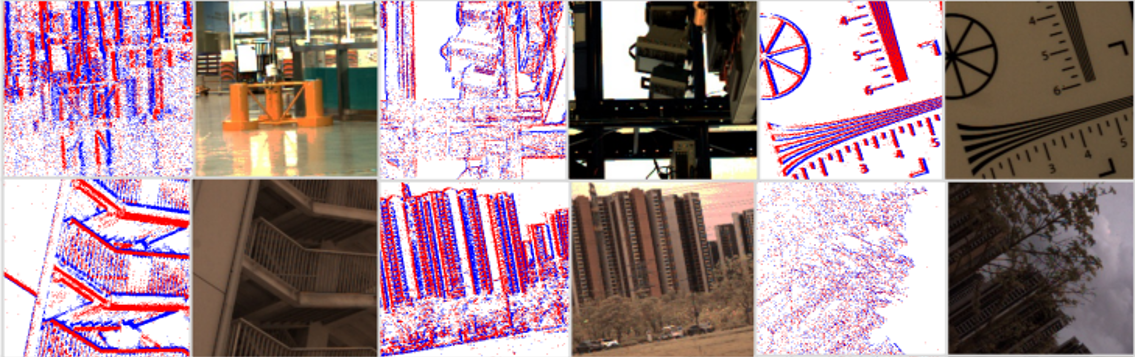}} & \begin{tabular}{l}
$346 \times 260$
\end{tabular} & \begin{tabular}{l}
5,000 pairs \\
$\approx$ 20 categories \\
16 sequences \\
\end{tabular} & \begin{tabular}{l}
Real-world \\
DAVIS346C \\
Low-resolution \\
Daliy Scene \\
\href{https://github.com/yunfanLu/SEE}{Download}
\end{tabular} \\

\hline VisEvent~\cite{wang2023visevent} &\raisebox{-.5\height}{\includegraphics[width=5.0cm,height=1.9cm]{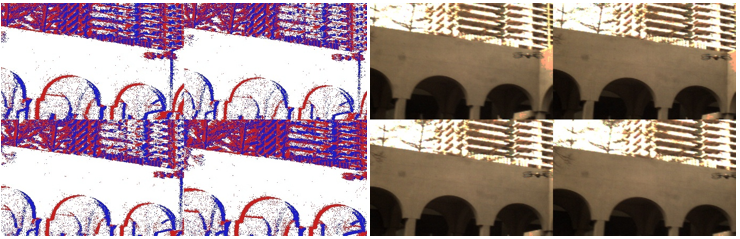}} & \begin{tabular}{l}
$346 \times 260$
\end{tabular} & \begin{tabular}{l}
30,000 pairs \\
$\approx$ 80 categories \\
820 sequences \\
\end{tabular} & \begin{tabular}{l}
Real-world \\
DAVIS346C \\
Low-resolution \\
Daliy Scene \\
\href{https://github.com/wangxiao5791509/VisEvent_SOT_Benchmark}{Download}
\end{tabular} \\

\hline CoeSot~\cite{tang2025revisiting} &\raisebox{-.5\height}{\includegraphics[width=5.0cm,height=1.9cm]{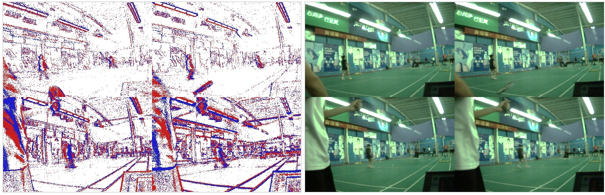}} & \begin{tabular}{l}
$346 \times 260$
\end{tabular} & \begin{tabular}{l}
30,000 pairs \\
$\approx$ 90 categories \\
1343 sequences \\
\end{tabular} & \begin{tabular}{l}
Real-world \\
DAVIS346C \\
Low-resolution \\
Daliy Scene \\
\href{https://github.com/Event-AHU/COESOT}{Download}
\end{tabular} \\

\hline DSEC~\cite{gehrig2021dsec} &\raisebox{-.5\height}{\includegraphics[width=5.0cm,height=1.9cm]{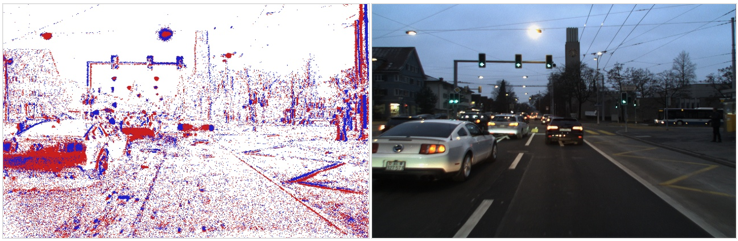}} & \begin{tabular}{l}
$640 \times 480$
\end{tabular} & \begin{tabular}{l}
20,000 pairs \\
$\approx$ 40 categories \\
53 sequences \\
\end{tabular} & \begin{tabular}{l}
Real-world \\
Prophesee Gen3.1\\
High-resolution \\
Driving Scene \\
\href{https://dsec.ifi.uzh.ch/dsec-datasets/download/}{Download}
\end{tabular} \\

\hline FEVD~\cite{kim2024frequency} &\raisebox{-.5\height}{\includegraphics[width=5.0cm,height=1.9cm]{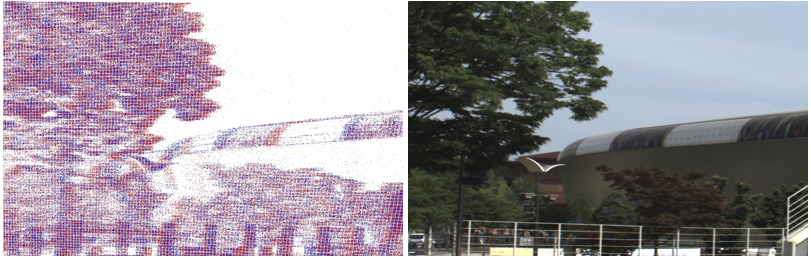}} & \begin{tabular}{l}
$1024 \times 768$
\end{tabular} & \begin{tabular}{l}
5,000 pairs \\
$\approx$ 20 categories \\
21 sequences \\
\end{tabular} & \begin{tabular}{l}
Real-world \\
Prophesee Gen4\\
High-resolution \\
Daliy Scene \\
\href{https://sites.google.com/view/fevd-cvpr2024}{Download}
\end{tabular} \\

\hline M3ED~\cite{Chaney_2023_CVPR} &\raisebox{-.5\height}{\includegraphics[width=5.0cm,height=1.9cm]{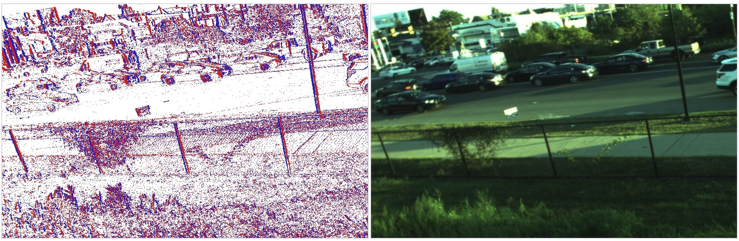}} & \begin{tabular}{l}
$1280 \times 720$
\end{tabular} & \begin{tabular}{l}
20,000 pairs \\
$\approx$ 40 categories \\
57 sequences \\
\end{tabular} & \begin{tabular}{l}
Real-world \\
Prophesee Gen4\\
High-resolution \\
Multiple Platforms \\
\href{https://m3ed.io/sequences/#car-1}{Download}
\end{tabular} \\

\hline HighREV~\cite{sun2023event} &\raisebox{-.5\height}{\includegraphics[width=5.0cm,height=1.9cm]{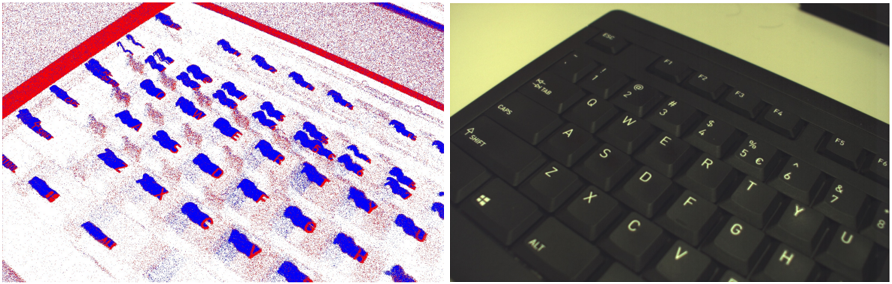}} & \begin{tabular}{l}
$1632 \times 1224$
\end{tabular} & \begin{tabular}{l}
10,000 pairs \\
$\approx$ 20 categories \\
25 sequences \\
\end{tabular} & \begin{tabular}{l}
Real-world \\
High-resolution \\
Multi-modality \\
Daliy Scene \\
\href{https://www.kaggle.com/datasets/lei0331/highrev-full/code}{Download}
\end{tabular} \\
\hline
\end{tabular}
\label{tab:7}
\vskip 0.5in
\end{table*}

\begin{table*}
\caption{The pretraining dataset configuration and data statistics for the \textbf{seven synthetic event-image datasets} used in our experiments.}
\centering
\renewcommand{\arraystretch}{1.10}
\setlength{\tabcolsep}{5.5pt}
\small
\begin{tabular}{l|l|l|l|l}
\hline Dataset & Illustration &Resolution & Statistics &Source\&Type \\
\hline SDSD~\cite{wang2021seeing} &\raisebox{-.5\height}{\includegraphics[width=5.0cm,height=1.9cm]{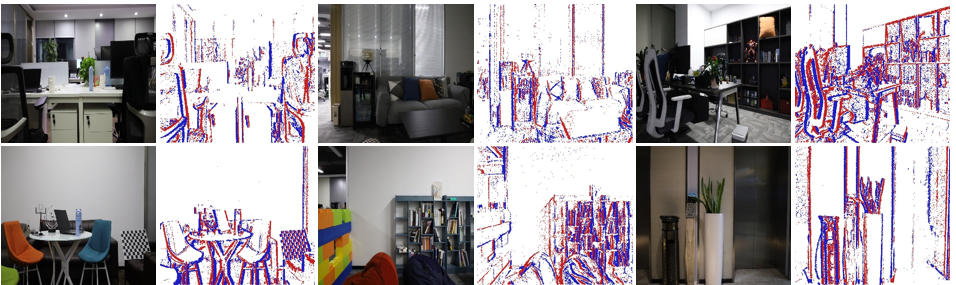}} & \begin{tabular}{l}
$346 \times 260$
\end{tabular} & \begin{tabular}{l}
20,000 pairs \\
$\approx$ 50 categories \\
150 sequences \\
\end{tabular} & \begin{tabular}{l}
VID2E Simulation \\
Low-resolution \\
Daliy Scene \\
\href{https://github.com/dvlab-research/SDSD}{Download}
\end{tabular} \\

\hline DAVIS17~\cite{Pont-Tuset_arXiv_2017} &\raisebox{-.5\height}{\includegraphics[width=5.0cm,height=1.9cm]{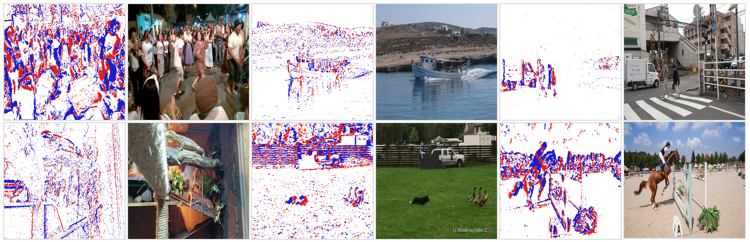}} & \begin{tabular}{l}
$346 \times 260$
\end{tabular} & \begin{tabular}{l}
20,000 pairs \\
$\approx$ 100 categories \\
90 sequences \\
\end{tabular} & \begin{tabular}{l}
VID2E Simulation \\
Low-resolution \\
Motion Scene \\
\href{https://github.com/BIT-Vision/ECOS?tab=readme-ov-file}{Download}
\end{tabular} \\

\hline DECD~\cite{rebecq2019high} &\raisebox{-.5\height}{\includegraphics[width=5.0cm,height=1.9cm]{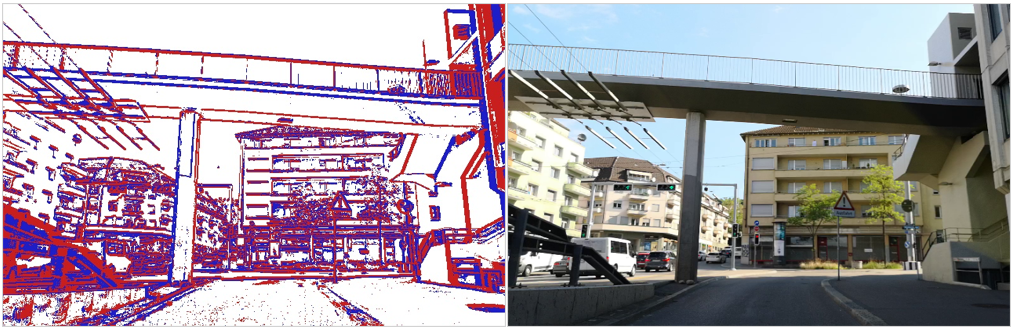}} & \begin{tabular}{l}
$640 \times 480$
\end{tabular} & \begin{tabular}{l}
40,000 pairs \\
$\approx$ 40 categories \\
120 sequences \\
\end{tabular} & \begin{tabular}{l}
VID2E Simulation \\
High-resolution \\
Driving Scene \\
\href{https://rpg.ifi.uzh.ch/E2VID.html}{Download}
\end{tabular} \\

\hline KITTI~\cite{geiger2013vision} &\raisebox{-.5\height}{\includegraphics[width=5.0cm,height=1.85cm]{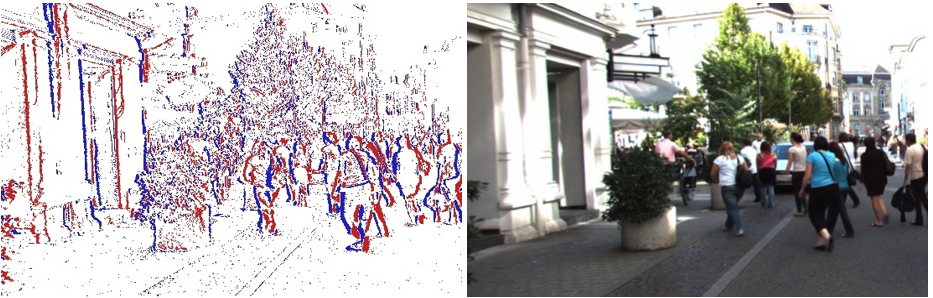}} & \begin{tabular}{l}
$1242 \times 375$
\end{tabular} & \begin{tabular}{l}
30,000 pairs \\
$\approx$ 40 categories \\
60 sequences \\
\end{tabular} & \begin{tabular}{l}
VID2E Simulation \\
High-resolution \\
Driving Scene \\
\href{https://www.cvlibs.net/datasets/kitti/}{Download}
\end{tabular} \\

\hline GoPro~\cite{nah2019ntire} &\raisebox{-.5\height}{\includegraphics[width=5.0cm,height=1.85cm]{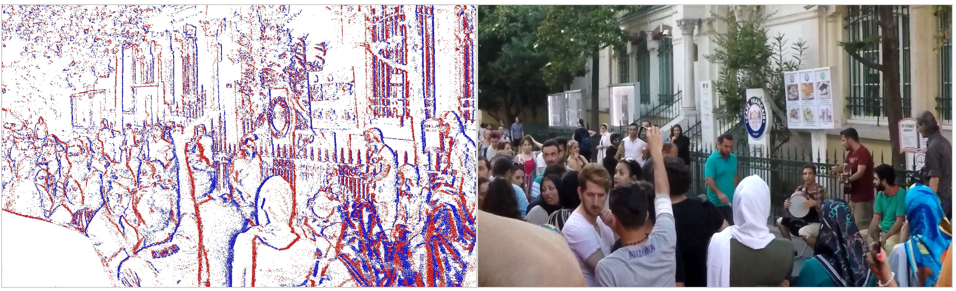}} & \begin{tabular}{l}
$1280 \times 720$\\
\end{tabular} & \begin{tabular}{l}
10,000 pairs \\
$\approx$ 30 categories \\
35 sequences \\
\end{tabular} & \begin{tabular}{l}
VID2E Simulation \\
High-resolution \\
Daliy Scene \\
\href{https://pkuml.org/resources/pku-ddd17-car.html}{Download}
\end{tabular} \\

\hline  Waymo~\cite{sun2020scalability} &\raisebox{-.5\height}{\includegraphics[width=5.0cm,height=1.85cm]{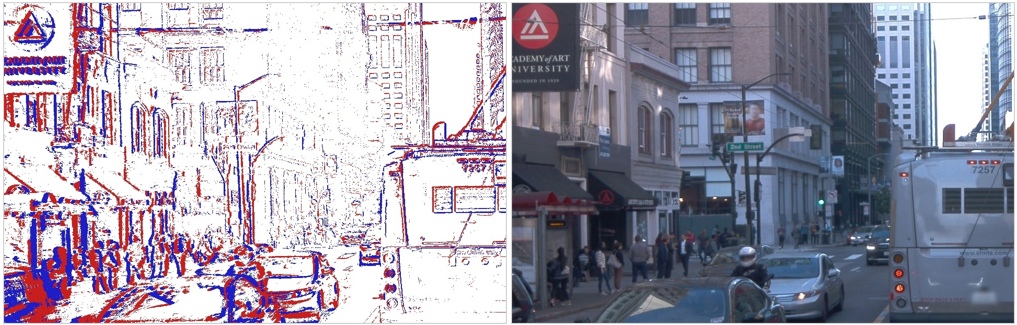}} & \begin{tabular}{l}
$1920 \times 1280$
\end{tabular} & \begin{tabular}{l}
50,000 pairs \\
$\approx$ 40 categories \\
147 sequences \\
\end{tabular} & \begin{tabular}{l}
VID2E Simulation \\
High-resolution \\
Driving Scene \\
\href{https://waymo.com/open/data/motion/}{Download}
\end{tabular} \\

\hline Cityscapes~\cite{cordts2016cityscapes} &\raisebox{-.5\height}{\includegraphics[width=5.0cm,height=1.85cm]{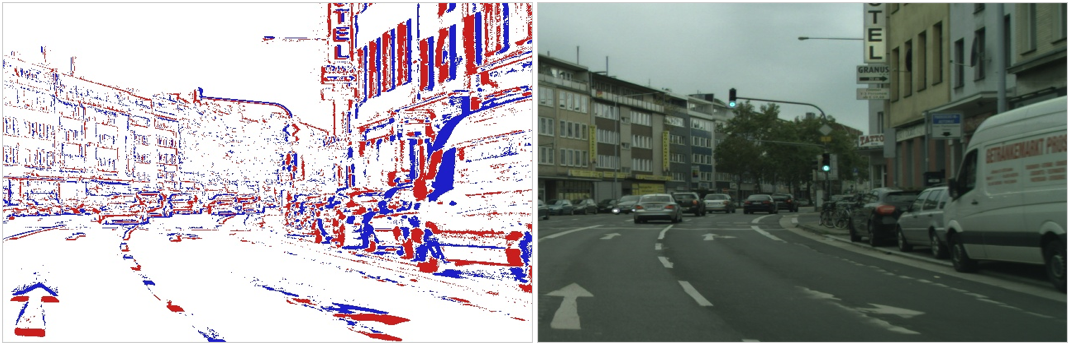}} & \begin{tabular}{l}
$2048 \times 1024$
\end{tabular} & \begin{tabular}{l}
200,000 pairs \\
$\approx$ 40 categories \\
10000 sequences \\
\end{tabular} & \begin{tabular}{l}
VID2E Simulation \\
High-resolution \\
Driving Scene \\
\href{https://www.cityscapes-dataset.com/downloads/}{Download}
\end{tabular} \\
\hline
\end{tabular}
\label{tab:8}
\vskip -0.05 in
\end{table*}

\subsection{Downstream Datasets}
\noindent \textbf{Semantic Segmentation.} Following prior works~\cite{liang2025efficient, zhao2025eseg, li2025know}, we evaluate event-based semantic segmentation on the DDD17-Seg~\cite{alonso2019ev} and DSEC-Semantic~\cite{sun2022ess} datasets.

\textbf{(i) DDD17-Seg}: DDD17-Seg~\cite{alonso2019ev} is a semantic segmentation extension of the DDD17~\cite{binas2017ddd17} dataset. Alonso and Murillo~\cite{alonso2019ev} overlay semantic masks on by leveraging co-registered gray-scale frames with event streams to synthesize approximate labels, which proved effective for training models that segment directly on event data. The dataset provides $15,950$ training and $3,890$ test samples, with semantic maps at $352 \times 200$ resolution. Each pixel is annotated with one of six classes: \emph{flat, background, object, vegetation, human, and vehicle}. \href{https://github.com/Shathe/Ev-SegNet}{Download}.

\textbf{(ii) DSEC-Semantic}: DSEC-Semantic~\cite{sun2022ess} is a semantic segmentation extension of the DSEC~\cite{gehrig2021dsec} dataset. Leveraging DSEC’s synchronized, high-resolution RGB images and event streams across diverse driving conditions, Sun et al.~\cite{sun2022ess} applied a pseudo-labeling procedure akin to DDD17-Seg\cite{alonso2019ev} to generate semantic masks for eleven sequences (11/53), yielding the DSEC-Semantic benchmark. The dataset provides $8,082$ training and $2,809$ test samples, with semantic maps at $640 \times 440$ resolution. Each pixel is annotated with one of eleven classes: \emph{ background, building, fence, person, pole, road, sidewalk, vegetation, car, wall, and traffic-sign}. \href{https://github.com/uzh-rpg/ess?tab=readme-ov-file}{Download}.

\vskip 0.05in
\noindent \textbf{Depth Estimation.} Following the setup of prior works~\cite{liu2024event,bartolomei2025depth}, we evaluate on the MVSEC-Depth~\cite{zhu2018multivehicle} and DSEC-Depth datasets~\cite{gehrig2021dsec} for event-based monocular depth estimation. 

\textbf{(i) MVSEC-Depth}: MVSEC-Depth is a depth estimation variant of the MVSEC~\cite{zhu2018multivehicle} dataset. The dataset provides events at a resolution $346 \times 260$ pixels from a stereo event camera consisting of two DAVIS346B sensors. The depth ground-truth is derived from a 16-line LiDAR using Lidar Odometry and Mapping (LOAM), yielding a total of $10,351$ training samples and $21,125$ testing samples. The test set is divided into a $5k$-sample daytime subset and three night-time subsets, each containing $5k$ samples. \href{https://daniilidis-group.github.io/mvsec/}{Download}.

\textbf{(ii) DSEC-Depth}: DSEC-Depth is a depth estimation variant of the DSEC~\cite{gehrig2021dsec} dataset. DSEC employs two Prophesee Gen3.1 event cameras in a stereo configuration. The disparity ground-truth is obtained using a 32-beam LiDAR, processed with a Lidar Inertial Odometry algorithm, and further filtered to remove outliers. We convert the disparity ground-truth to depth map based on the stereo setup parameters. The dataset provides $19,181$ training and $7,157$ test samples, with depth maps at $640 \times 480$ resolution. \href{https://dsec.ifi.uzh.ch/dsec-datasets/download/}{Download}.

\vskip 0.05in
\noindent \textbf{Optical Flow Estimation.}  Following the setup of prior works~\cite{yang2023event,yang2024event}, we evaluate event-based optical flow estimation on the MVSEC-Flow~\cite{zhu2018ev} dataset. \textbf{MVSEC-Flow} is an optical flow estimation variant of the MVSEC~\cite{zhu2018multivehicle} dataset. MVSEC employs two DAVIS346B event cameras in a stereo configuration. MVSEC-Flow provides per-camera poses and depth maps for each event camera, which were used to generate ground truth optical flow. In this work, we use \emph{outdoor$\_$day2} sequence for training ($26,677$ samples), \emph{indoor$\_$flying1, indoor$\_$flying2, indoor$\_$flying3} sequences for testing ($7,775$ samples). \href{https://github.com/daniilidis-group/EV-FlowNet}{Download}.

\subsection{Downstream Fine-tuning} 
\begin{itemize}
  \item \textbf{Experimental Setup.} The details of the fine-tuning procedure are outlined in Table~\ref{tab:fine-tune}.

  \item \textbf{Data Augmentation.} No data augmentation strategies are applied during fine-tuning on downstream tasks. 
  
  \item \textbf{Linear Probing.} The pretrained event feature encoder is frozen with a trainable pixel-wise task head which is trained for 20 epochs, setting the initial learning rate at $5 \times 10^{-4}$, with a weight decay of $1 \times 10^{-4}$. 

  \item \textbf{Few-shot Fine-tuning.} In few-shot fine-tuning, we subsample the training split of each downstream dataset to obtain $1\%$, $5\%$, $10\%$, or $20\%$ annotated scans, generated via fixed-interval sampling over the full training sequences, such as $100$, $20$, $10$, $5$.

\end{itemize}

\begin{table*}[htbp]
\caption{\textbf{Experimental setup for fine-tuning downstream tasks.} \textbf{lr} denotes learning rate. All configurations are based on the ViT-L encoder. Apart from batch size, which depends on model scale, all other settings remain identical across experiments.}
\vskip -0.05in
\centering
\renewcommand{\arraystretch}{1.0}
\setlength{\tabcolsep}{7.0pt}
\footnotesize
\begin{tabular}{c| c | c | c | c | c }
    \midrule[1.1pt]
    \multirow{2}{*}{\textbf{Dataset}}  &\multicolumn{2}{c|}{\textbf{Semantic Segmentation}} &\multicolumn{2}{c|}{\textbf{Depth Estimation}} &\multicolumn{1}{c}{\textbf{Flow Estimation}}  \\
    \cline{2-6}
    \multicolumn{1}{c|}{}  &DDD17-Seg &DSEC-Semantic &MVSEC-Depth &DSEC-Depth &MVSEC-Flow  \\
    \midrule[1.1pt]
    optimizer & AdamW & AdamW & AdamW & AdamW & AdamW \\
    \hline encoder lr & $2 \times 10^{-6}$ & $2 \times 10^{-6}$ &$2 \times 10^{-6}$ &$1 \times 10^{-6}$ &$1 \times 10^{-6}$ \\
    \hline decoder lr & $5 \times 10^{-6}$ & $4 \times 10^{-6}$ &$4 \times 10^{-6}$ &$2 \times 10^{-6}$ &$2 \times 10^{-6}$ \\
    \hline weight decay &$1 \times 10^{-4}$ &$1 \times 10^{-4}$ &$1 \times 10^{-4}$ &$1 \times 10^{-4}$ &$1 \times 10^{-4}$ \\
    \hline batch size &40 &12 &40  &12 &24  \\
    \hline epochs & 20 &30 &30 &30 &20 \\
    \hline lr scheduler &exponential &exponential &exponential &exponential &exponential \\
    \hline scheduler gamma &0.9 &0.9 &0.9 &0.9 &0.9 \\
    \hline scheduler epoch &5 &5 &5 &5 &5 \\
    \hline gradient clipping norm &0.1 &0.1 &0.1 &0.1 &0.1 \\
    \bottomrule[1.1pt]
\end{tabular}
\label{tab:fine-tune}
\vskip -0.1 in
\end{table*}

\section{More Quantitative Results}
\label{appendix:quantitative}

\begin{table}
\caption{The per-class segmentation results of our methods on the DDD17-Seg dataset. Scores reported are IoUs in percentage.}
\vskip -0.05in
\centering
\renewcommand{\arraystretch}{1.20}
\setlength{\tabcolsep}{3.0pt}
\footnotesize
\begin{tabular}{c|c|cccccc|c}
\hline Event Model &\textbf{\rotatebox{90}{mIoU}} & \rotatebox{90}{flat} &\rotatebox{90}{background}  &\rotatebox{90}{object} &\rotatebox{90}{vegetation} &\rotatebox{90}{human} &\rotatebox{90}{vehicle} &\textbf{\rotatebox{90}{Acc}} \\
\rowcolor{blue!6} \multicolumn{9}{c}{\textbf{Linear Probing}}   \\
\hline ViT-S/16 &55.64 &79.61 &91.18 &15.90 &57.51 &22.02 &67.72 &91.27 \\
\hline ViT-B/16 &57.87 &79.92 &91.24 &15.87 &58.04 &34.97 &67.05 &91.31 \\
\hline ViT-L/16 &60.30 &81.03 &91.49 &18.83 &57.21 &44.18 &68.95 &91.83 \\
\rowcolor{gray!10} \multicolumn{9}{c}{\textbf{Fine-Tuning (1$\%$)}}   \\
\hline ViT-S/16 &53.87 &78.61 &90.06 &10.03 &54.59 &25.18 &64.63 &90.41  \\
\hline ViT-B/16 &57.23 &79.51 &91.03 &15.46 &57.53 &34.87 &66.93 &91.12 \\
\hline ViT-L/16 &59.23 &82.34 &92.24 &18.26 &61.68 &34.01 &69.37 &91.68 \\
\rowcolor{gray!10} \multicolumn{9}{c}{\textbf{Fine-Tuning (5$\%$)}}   \\
\hline ViT-S/16 &54.36 &78.96 &90.27 &10.38 &54.92 &27.26 &64.40 &90.62  \\
\hline ViT-B/16 &59.54 &80.25 &91.65 &15.24 &59.21 &44.38 &65.80 &91.65 \\
\hline ViT-L/16 &62.52 &81.96 &91.97 &19.31 &61.49 &50.77 &69.63 &92.12 \\
\rowcolor{gray!10} \multicolumn{9}{c}{\textbf{Fine-Tuning (10$\%$)}}   \\
\hline ViT-S/16 &57.29 &79.52 &91.16 &12.24 &59.26 &39.77 &64.72 &91.34 \\
\hline ViT-B/16 &61.45 &82.37 &91.69 &20.14 &60.69 &45.16 &68.46 &91.72 \\
\hline ViT-L/16 &63.71 &82.23 &92.12 &23.75 &59.84 &51.80 &72.95 &92.13 \\
\rowcolor{gray!10} \multicolumn{9}{c}{\textbf{Fine-Tuning (20$\%$)}}   \\
\hline ViT-S/16 &58.37 &79.93 &91.55 &13.02 &58.93 &41.81 &66.27 &91.63 \\
\hline ViT-B/16 &62.06 &82.74 &92.05 &18.72 &61.65 &49.79 &69.60 &92.24 \\
\hline ViT-L/16 &64.43 &83.10 &92.23 &23.17 &62.62 &54.13 &71.35 &92.44 \\
\rowcolor{gray!10} \multicolumn{9}{c}{\textbf{Fine-Tuning (100$\%$)}}   \\
\hline ViT-S/16 &59.64 &80.68 &91.27 &17.58 &58.88 &43.71 &65.73 &91.39 \\
\hline ViT-B/16 &62.81 &82.95 &92.00 &18.79 &61.71 &51.43 &69.98  &92.21 \\
\hline ViT-L/16 &65.09 &83.73 &92.34 &23.10 &62.61 &56.43 &72.26 &92.62 \\
\hline
\end{tabular}
\label{tab:per-class-segment-ddd17}
\vskip -0.05in
\end{table}

\subsection{More Detailed Comparisons}
We report the complete results (i.e., the class-wise IoU scores, optical flow/depth metrics) for the \textbf{inear probing} and \textbf{downstream fine-tuning tasks} outlined in the main paper. Specifically, the detailed performance metrics on the DDD17-Seg, DSEC-Semantic, MVSEC-Depth, DSEC-Depth, and MVSEC-Flow datasets are shown in Table~\ref{tab:per-class-segment-ddd17}, Table~\ref{tab:per-class-segment-dsec}, Table~\ref{tab:per-sequence-depth} and Table~\ref{tab:per-sequence-flow}, respectively. These results comprehensively evaluate the model's performance across a variety of dense perception tasks.

\begin{table}
\caption{The optical flow results of our methods on the MVSEC-Flow dataset.}
\vskip -0.15in
\centering
\renewcommand{\arraystretch}{1.15}
\setlength{\tabcolsep}{4.0pt}
\footnotesize
\begin{tabular}{c | c c |c c | c c}
    \midrule[1.1pt]
    \multirow{2}{*}{Event Model}   &\multicolumn{2}{c|}{indoor flying1} &\multicolumn{2}{c|}{indoor flying2} &\multicolumn{2}{c}{indoor flying3}  \\
    \cline{2-7}
    \multicolumn{1}{c|}{} &EPE $\downarrow$ &Out$\downarrow$ &EPE $\downarrow$ &Out$\downarrow$ &EPE $\downarrow$ &Out$\downarrow$ \\
    \hline
    ViT-S/16   &0.29 &0.03 &0.38 &0.001 &0.40 &0.001 \\
    \hline
    ViT-B/16   &0.28 &0.03 &0.38 &0.001 &0.39 &0.001 \\
    \hline
    ViT-L/16   &0.27 &0.03 &0.37 &0.001 &0.39 &0.001 \\
    \hline
\end{tabular}
\label{tab:per-sequence-flow}
\vskip -0.15in
\end{table}


\begin{table*}
\caption{The per-class segmentation results of our methods on the DSEC-Semantic dataset. Scores reported are IoUs in percentage.}
\centering
\renewcommand{\arraystretch}{1.15}
\setlength{\tabcolsep}{6.0pt}
\footnotesize
\begin{tabular}{c|c|ccccccccccc|c}
\hline Event Model &\textbf{\rotatebox{90}{mIoU}} &\rotatebox{90}{background} & \rotatebox{90}{building} &\rotatebox{90}{fence} &\rotatebox{90}{person} &\rotatebox{90}{pole} &\rotatebox{90}{road} &\rotatebox{90}{sidewalk} & \rotatebox{90}{vegetation} &\rotatebox{90}{car} &\rotatebox{90}{wall} &\rotatebox{90}{traffic-sign} &\textbf{\rotatebox{90}{Acc}} \\
\rowcolor{blue!6} \multicolumn{14}{c}{\textbf{Linear Probing}}   \\
\hline ViT-S/16 &55.46 &92.81 &81.88 &17.45 &15.67 &24.98 &93.20 &68.12 &78.83 &77.72 &30.79 &43.86  &90.12 \\
\hline ViT-B/16 &58.42 &93.46 &83.52 &23.88 &16.69 &27.85 &93.72 &69.27 &80.38 &80.13 &43.06 &43.25 &91.44 \\
\hline ViT-L/16 &61.29 &93.91 &85.09 &27.66 &27.37 &33.58 &93.34 &70.94 &82.37 &82.27 &41.82 &48.66 &91.69 \\
\rowcolor{gray!10} \multicolumn{14}{c}{\textbf{Fine-Tuning (1$\%$)}}   \\
\hline ViT-S/16 &52.97 &92.35 &81.36 &18.04 &7.93 &18.77 &92.55 &60.54 &78.50 &76.22 &22.25 &37.06 &89.56 \\
\hline ViT-B/16 &54.37 &93.04 &82.55 &14.09 &16.14 &26.06 &93.55 &65.17 &80.79 &79.52 &12.34 &41.40 &90.14 \\
\hline ViT-L/16 &59.73 &92.96 &82.55 &21.88 &20.30 &27.87 &93.34 &69.36 &80.68 &80.22 &40.60 &47.24 &90.73 \\
\rowcolor{gray!10} \multicolumn{14}{c}{\textbf{Fine-Tuning (5$\%$)}}   \\
\hline ViT-S/16 &56.55 &93.01 &82.08 &19.92 &18.34 &19.48 &93.15 &66.34 &79.36 &79.22 &33.71 &46.98 &90.78 \\
\hline ViT-B/16 &62.87 &93.66 &84.91 &22.14 &33.63 &31.05 &93.91 &71.36 &81.84 &82.27 &44.28 &49.59 &91.52 \\
\hline ViT-L/16 &68.03 &94.68 &86.68 &31.01 &52.67 &39.86 &94.77 &74.13 &84.08 &83.89 &49.93 &58.39 &92.83 \\
\rowcolor{gray!10} \multicolumn{14}{c}{\textbf{Fine-Tuning (10$\%$)}}   \\
\hline ViT-S/16 &58.96 &93.56 &84.73 &21.88 &19.25 &22.50 &93.26 &68.90 &79.52 &78.61 &42.29 &42.48 &91.25 \\
\hline ViT-B/16 &63.88 &93.59 &85.06 &23.05 &38.58 &31.72 &94.05 &71.48 &81.85 &82.38 &48.44 &49.10 &91.73 \\
\hline ViT-L/16 &68.51  &94.52 &86.76 &26.70 &51.15 &44.29 &94.85 &75.24 &84.49 &84.92 &47.71 &62.72 &92.92\\
\rowcolor{gray!10} \multicolumn{14}{c}{\textbf{Fine-Tuning (20$\%$)}}   \\
\hline ViT-S/16 &60.20 &93.32 &83.99 &22.61 &27.38 &29.98 &93.07 &69.40 &79.62 &80.29 &33.19 &48.69 &91.62 \\
\hline ViT-B/16 &64.15 &93.45 &84.93 &24.27 &40.01 &32.64 &93.91 &71.77 &82.13 &82.55 &50.67 &50.30 &91.78 \\
\hline ViT-L/16 &69.25 &94.63 &86.77 &26.98  &51.24 &44.27 &94.90 &75.45 &84.74 &84.93 &47.90 &62.95 &92.98 \\
\rowcolor{gray!10} \multicolumn{14}{c}{\textbf{Fine-Tuning (100$\%$)}}   \\
\hline ViT-S/16 &61.12 &93.16 &83.16 &26.28 &34.11 &30.15 &93.12 &68.21 &80.10 &80.04 &34.89 &49.03 &90.76 \\
\hline ViT-B/16 &64.93 &93.43 &85.17 &23.69 &42.63 &34.14 &94.08 &72.84 &82.28 &83.13 &51.86 &50.96 &92.00 \\
\hline ViT-L/16 &69.65 &94.54 &86.71 &26.88 &55.63 &45.53 &95.13 &76.64  &83.94 &86.13 &52.53 &62.44 &93.10 \\
\hline
\end{tabular}
\label{tab:per-class-segment-dsec}
\vskip -0.05in
\end{table*}

\begin{table*}
\caption{The depth results of our methods on the MVSEC-Depth and DSEC-Depth datasets.}
\centering
\renewcommand{\arraystretch}{1.20}
\setlength{\tabcolsep}{5.5pt}
\footnotesize
\begin{tabular}{c| c | c c c c c c | c c c c c c}
    \hline
    \multirow{2}{*}{Event Model} &\multirow{2}{*}{Metric}  &\multicolumn{6}{c|}{\textbf{MVSEC-Depth}} &\multicolumn{6}{c}{\textbf{DSEC-Depth}}  \\
    \cline{3-14}
    \multicolumn{1}{c|}{} &  & LP & 1\% & 5\% & 10\% & 20\% & Full & LP & 1\% & 5\% & 10\% & 20\% & Full \\
    \hline
    \multirow{2}{*}{ViT-S/16} &$\delta_1$ $\uparrow$ &0.529 &0.526 &0.531 &0.542 &0.560 &0.577 &0.798 &0.795 &0.804 &0.811 &0.816 &0.824  \\
    \cline{2-14}
    &RMSE$\downarrow$ &6.756 &6.930 &6.712 &6.477 &6.352 &6.145 &4.861 &4.983 &4.751 &4.728 &4.694 &4.564 \\
    \hline
    \multirow{2}{*}{ViT-B/16} &$\delta_1$ $\uparrow$ &0.571 &0.561 &0.574 &0.587 &0.591 &0.594 &0.845 &0.839 &0.856 &0.863 &0.867 &0.872 \\
    \cline{2-14}
    &RMSE$\downarrow$ &6.392 &6.546 &6.339 &6.012 &5.908 &5.891 &4.352 &4.471 &4.264 &4.192 &4.154 &4.032 \\
    \hline
    \multirow{2}{*}{ViT-L/16} &$\delta_1$ $\uparrow$ &0.597 &0.592 &0.601 &0.612 &0.619 &0.625 &0.881 &0.856 &0.883 &0.892 &0.893 &0.896 \\
    \cline{2-14}
    &RMSE$\downarrow$ &5.884 &5.975 &5.855 &5.724 &5.673 &5.554 &3.857 &3.984 &3.841 &3.759 &3.723 &3.694 \\
    \hline
\end{tabular}
\label{tab:per-sequence-depth}
\vskip -0.05in
\end{table*}

\subsection{More Detailed Ablations}

\begin{table}[thbp]
\caption{Ablative study results of different event aggregation methods.}
\vskip -0.05in
\centering
\renewcommand{\arraystretch}{1.20}
\setlength{\tabcolsep}{3.5pt}
\footnotesize           
\begin{tabular}{c | c  c |c c | cc}
   \toprule
    \multirow{2}{*}{\textbf{Event Input}} &\multicolumn{2}{c|}{\textbf{DDD17-Seg}} &\multicolumn{2}{c|}{\textbf{DSEC-Depth}} &\multicolumn{2}{c}{\textbf{MVSEC-Flow}}\\
    \cline{2-7}
    \multicolumn{1}{c|}{} &Acc$\uparrow$ &mIoU$\uparrow$ &$\delta_1$ $\uparrow$  &RMSE $\downarrow$ &EPE$\downarrow$ &Out$\downarrow$ \\
    \hline
    Color Frame  &90.76 &56.37  &0.784 &5.306  &1.107 &6.720  \\
    \hline
    E2VID &89.25  &55.72  &0.809 &4.928  &0.852  &3.294  \\
    \hline
    \textbf{Event Volume}  &91.39  &59.64  &0.824  &4.564  &0.356  &0.094  \\
    \hline
\end{tabular}
\label{tab:ablation-aggregation}
\end{table}

\noindent \textbf{Event Aggregations.} In our main study, we aggregate the event stream as a three-dimensional volume (voxel grid) to interface cleanly with vision foundation models. Here, we  additionally evaluate alternative renderings, including color-like frames~\cite{wang2023visevent} and E2VID reconstructions~\cite{rebecq2019high}. For a fair comparison, the event representation is held fixed across pretraining and downstream fine-tuning, and all experiments use a ViT-S encoder. As reported in Table~\ref{tab:ablation-aggregation}, the volumetric encoding delivers the strongest overall performance, indicating that explicit spatio-temporal discretization provides a more effective inductive bias for pretraining than image-like aggregations or reconstructed intensities.

\begin{table}[thbp]
\caption{Ablative study results of different time bins for event volume aggregation.}
\vskip -0.05in
\centering
\renewcommand{\arraystretch}{1.2}
\setlength{\tabcolsep}{3.5pt}
\footnotesize           
\begin{tabular}{c | c  c |c c | cc}
   \toprule
    \multirow{2}{*}{\textbf{Time Bin}} &\multicolumn{2}{c|}{\textbf{DDD17-Seg}} &\multicolumn{2}{c|}{\textbf{DSEC-Depth}} &\multicolumn{2}{c}{\textbf{MVSEC-Flow}}\\
    \cline{2-7}
    \multicolumn{1}{c|}{} &Acc$\uparrow$ &mIoU$\uparrow$ &$\delta_1$ $\uparrow$  &RMSE $\downarrow$ &EPE$\downarrow$ &Out$\downarrow$ \\
    \hline
    $B=1$  &91.07  &58.43  &0.819  &4.736  &0.365  &0.104  \\
    \hline
    $\bm{B=3}$ &91.39  &59.64  &0.824  &4.564  &0.356  &0.094  \\
    \hline
    $B=5$  &91.20  &59.22  &0.822  &4.613  &0.359  &0.095  \\
    \hline
\end{tabular}
\label{tab:ablation-time-bins}
\vskip -0.05in
\end{table}

\begin{table}[thbp]
\caption{Ablative study results of different density thresholds for activation mask constraint.}
\vskip -0.05in
\centering
\renewcommand{\arraystretch}{1.2}
\setlength{\tabcolsep}{3.5pt}
\footnotesize           
\begin{tabular}{c | c  c |c c | cc}
   \toprule
    \textbf{Density} &\multicolumn{2}{c|}{\textbf{DDD17-Seg}} &\multicolumn{2}{c|}{\textbf{DSEC-Depth}} &\multicolumn{2}{c}{\textbf{MVSEC-Flow}}\\
    \cline{2-7}
    \multicolumn{1}{c|}{\textbf{Threshold}} &Acc$\uparrow$ &mIoU$\uparrow$ &$\delta_1$ $\uparrow$  &RMSE $\downarrow$ &EPE$\downarrow$ &Out$\downarrow$ \\
    \hline
    $\tau = 32$  &91.33  &59.51  &0.823  &4.538  &0.362  &0.095  \\
    \hline
    $\bm{\tau = 64}$  &91.39  &59.64  &0.824  &4.564  &0.356  &0.094  \\
    \hline
    $\tau = 128$  &91.25  &59.32  &0.821  &4.640  &0.367  &0.097  \\
    \hline
\end{tabular}
\label{tab:ablation-density-threshold}
\vskip -0.05in
\end{table}

\begin{table}[thbp]
\caption{Ablative study results of different distillation objectives across granularities. CL denotes the contrastive loss.}
\vskip -0.05in
\centering
\renewcommand{\arraystretch}{1.2}
\setlength{\tabcolsep}{2.0pt}
\footnotesize           
\begin{tabular}{c | c  c |c c | cc}
   \toprule
    \textbf{Alignment} &\multicolumn{2}{c|}{\textbf{DDD17-Seg}} &\multicolumn{2}{c|}{\textbf{DSEC-Depth}} &\multicolumn{2}{c}{\textbf{MVSEC-Flow}}\\
    \cline{2-7}
    \multicolumn{1}{c|}{\textbf{Objective}} &Acc$\uparrow$ &mIoU$\uparrow$ &$\delta_1$ $\uparrow$  &RMSE $\downarrow$ &EPE$\downarrow$ &Out$\downarrow$ \\
    \hline
    patch-level (L1)  &90.65  &56.06  &0.785  &4.990  &0.367  &0.098  \\
    \hline
    superpixel-level (L1)  &90.88  &56.36  &0.790  &4.937  &0.384  &0.106  \\
    \hline
    superpixel-level (CL)  &90.92  &56.72  &0.782  &5.031  &0.435  &0.120  \\
    \hline
    \textbf{ours}  &91.39  &59.64  &0.824  &4.564  &0.356  &0.094 \\
    \hline
\end{tabular}
\label{tab:ablation-granularities}
\vskip -0.05in
\end{table}

\begin{table}[thbp]
\caption{Ablative study results of multi-scale distillation. }
\vskip -0.05in
\centering
\renewcommand{\arraystretch}{1.2}
\setlength{\tabcolsep}{4.0pt}
\footnotesize           
\begin{tabular}{c | c  c |c c | cc}
   \toprule
    \textbf{Alignment} &\multicolumn{2}{c|}{\textbf{DDD17-Seg}} &\multicolumn{2}{c|}{\textbf{DSEC-Depth}} &\multicolumn{2}{c}{\textbf{MVSEC-Flow}}\\
    \cline{2-7}
    \multicolumn{1}{c|}{\textbf{Objective}} &Acc$\uparrow$ &mIoU$\uparrow$ &$\delta_1$ $\uparrow$  &RMSE $\downarrow$ &EPE$\downarrow$ &Out$\downarrow$ \\
    \hline
    multi-scale  &91.07  &58.83  &0.816  &4.831  &0.377  &0.102  \\
    \hline
    \textbf{single-scale}  &91.39  &59.64  &0.824  &4.564  &0.356  &0.094 \\
    \hline
\end{tabular}
\label{tab:ablation-multi-scale}
\vskip -0.1in
\end{table}

\vskip 0.05in
\noindent \textbf{Hyper Parameters.} To enable cross-modal distillation, we encode event streams as a multi-channel volume/voxel grid compatible with vision foundation models and introduce an activation mask to suppress spurious event–image alignment during pretraining. 
We ablate two hyperparameters, the number of time bins $B$ for volume aggregation, which controls temporal granularity, and the density threshold $\tau$ for the activation mask, which trades coverage for noise suppression.  Unless otherwise specified, all comparisons use a ViT-S encoder. Results in Tables~\ref{tab:ablation-time-bins} and \ref{tab:ablation-density-threshold} identify the optimal configuration.

\vskip 0.05in
\noindent \textbf{Superpixel Alignment.} For cross-modal distillation, we formulate a hierarchical objective comprising patch-level supervision (our baseline) and structure-level supervision (our highlight). Here, we further examine a superpixel-level variant. Following OpenESS~\cite{kong2024openess}, we partition each image into 100 SAM-derived~\cite{kirillov2023segment} superpixels and compare two formulations: (i) an L1 regression loss on superpixel-aggregated features, and (ii) a contrastive objective inspired by image-point cloud distillation~\cite{sautier2022image} that enforces intra-superpixel compactness and inter-superpixel separability. Unless otherwise specified, all comparisons use a ViT-S encoder. Results in Table~\ref{tab:ablation-granularities} reveal that superpixel-level alignment underperforms, due to semantically ambiguous groupings (e.g., boundary leakage and region fragmentation) that is consistent with our overall analysis.

\begin{figure}
    \centering
    \includegraphics[clip, width=0.45\textwidth]{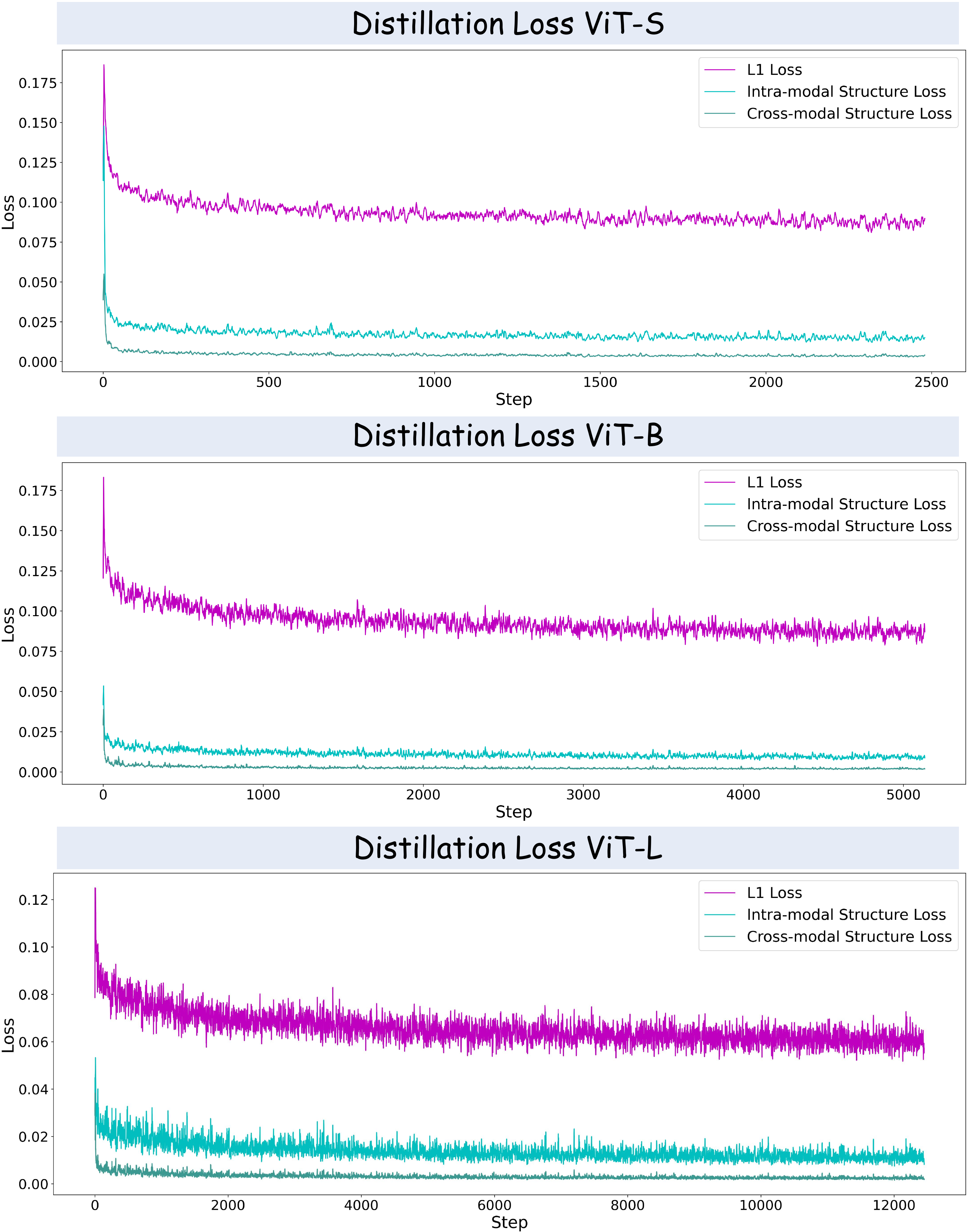}
    \caption{\textbf{Cross-modal distillation loss during pretraining} of our event-based ViT-S, ViT-B, and ViT-L feature encoders.}
    \label{fig:loss}
    \vspace{-0.5cm}
\end{figure}

\vskip 0.05in
\noindent \textbf{Multi-scale Distillation.} For cross-modal distillation, our main study aligns only the terminal features of the encoder. Here, we additionally assess a multi-scale alignment scheme. All comparisons use a ViT-S encoder. Specifically, we align intermediate activations from layers 3, 6, 9, and 12 to their event counterparts with equal loss weights. Results in Table~\ref{tab:ablation-multi-scale} show that multi-scale alignment underperforms, likely because intermediate representations possess weak and unstable semantics and thus exacerbate the event-image modality gap.

\subsection{Pretraining Loss}
The pretraining losses are depicted in Figure~\ref{fig:loss}.

\subsection{Computational Efficiency}
The computational efficiency analysis is shown in Table~\ref{tab:efficiency}.

\begin{table}[thbp]
\caption{\textbf{Computational efficiency} of our downstream task models, setting an input event volume resolution of $480 \times 640$.}
\vskip -0.05in
\centering
\renewcommand{\arraystretch}{1.20}
\setlength{\tabcolsep}{2.0pt}
\footnotesize           
\begin{tabular}{c | c  c |c c | cc}
   \toprule
    \textbf{} &\multicolumn{2}{c|}{\textbf{Segment Model}} &\multicolumn{2}{c|}{\textbf{Depth Model}} &\multicolumn{2}{c}{\textbf{Flow Model}}\\
    \cline{2-7}
    \multicolumn{1}{c|}{\textbf{}} &MParams &GFLOP &MParams &GFLOPs &MParams &GFLOPs \\
    \hline
    ViT-S  &28.23  &243.67  &18.92  &61.23  &40.95  &349.42  \\
    \hline
    ViT-B  &76.74  &363.82  &74.98  &231.93  &135.62  &962.71 \\
    \hline
    ViT-L  &239.09  &758.17  &257.29  &853.16  &485.27  &3369.48 \\
    \hline
\end{tabular}
\label{tab:efficiency}
\vskip -0.1in
\end{table}

\section{More Qualitative Results}
\label{appendix:qualitative}

\begin{figure*}
    \centering
    \includegraphics[clip, width=0.98\textwidth]{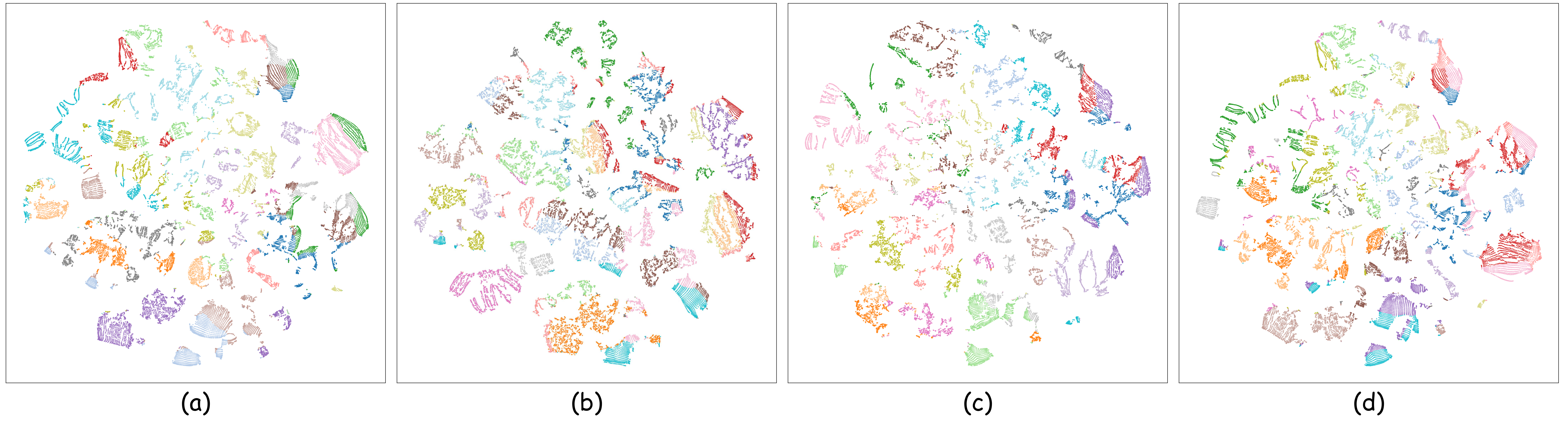}
    \caption{\textbf{T-SNE plots of learned event features.} We sample 20,000 event feature vectors from the DSEC dataset~\cite{gehrig2021dsec}. We show features from \textbf{(a)} images with pretrained DIVOv3-L;  \textbf{(b)} event volume with pretrained DIVOv3-L; \textbf{(c)} event volume with DIVOv3-L after patch-level distillation; \textbf{(d)} event volume with DIVOv3-L using our distillation method.}
    \label{fig:t-SNE}
    \vspace{-0.2cm}
\end{figure*}

\subsection{Representation Visualization}
\noindent \textbf{Statistical Analysis.}As shown in Figure~\ref{fig:t-SNE}, t-SNE plots of feature vectors from the DSEC dataset~\cite{gehrig2021dsec} highlight the performance of various models and distillation methods. The pretrained DINOv3-L on images shows strong clustering with some overlap, indicating effective feature learning but room for finer distinctions. Event volume with pretrained DINOv3-L shows greater dispersion, highlighting challenges in capturing event-specific features and temporal dynamics. Patch-level distillation improves feature separation, resulting in more compact clusters. Our distillation method achieves the most distinct and well-separated clusters, closely matching the pretrained DINOv3-L while better capturing event-specific features.

\begin{figure*}
    \centering
    \includegraphics[clip, width=0.98\textwidth]{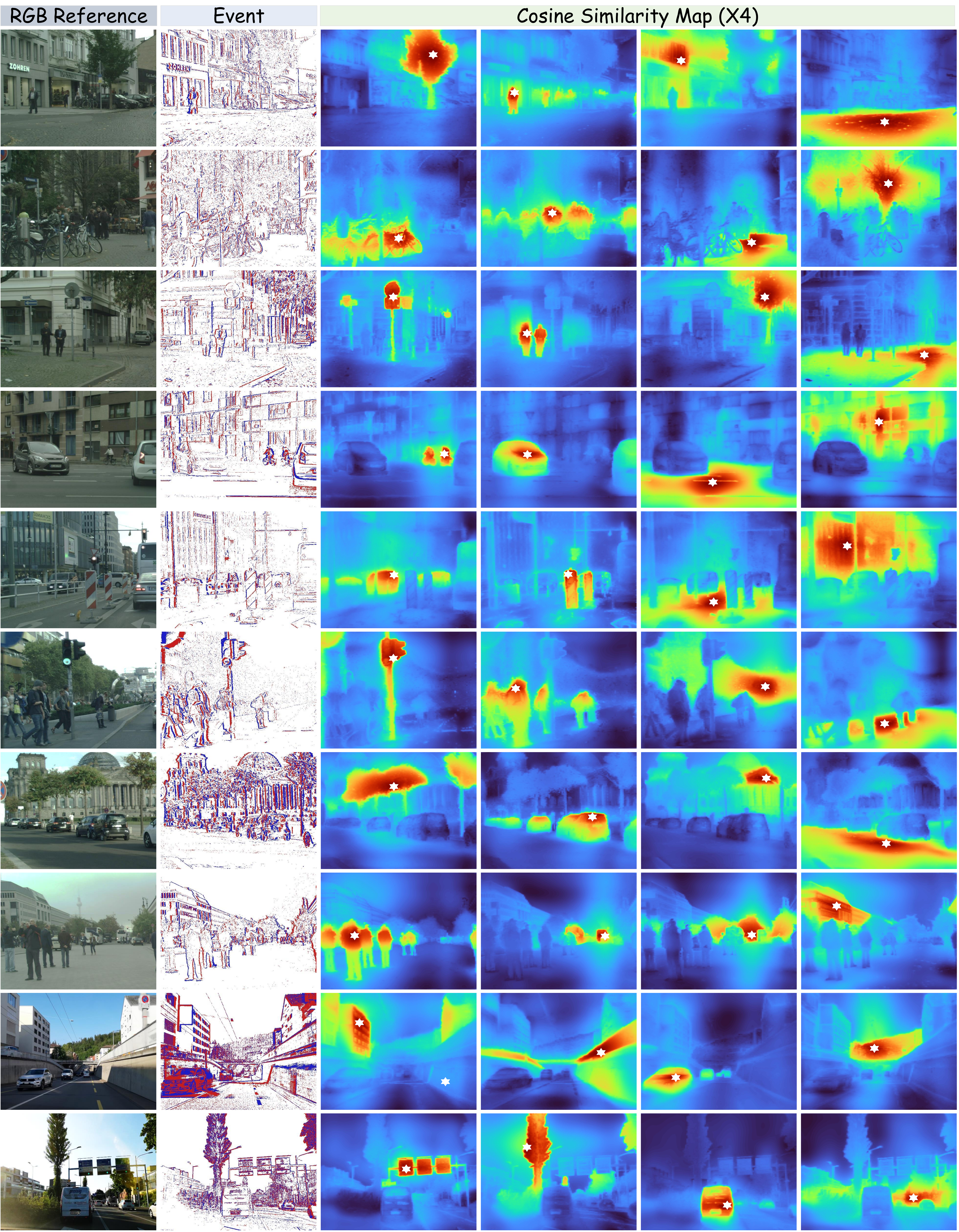}
    \caption{\textbf{The learned fine-grained event features (1/2)} of our method are primarily presented through cosine similarity maps, with key points anchored at the distinct white stars. Best viewed in color.}
    \label{fig:feature-v1}
\end{figure*}

\begin{figure*}
    \centering
    \includegraphics[clip, width=0.98\textwidth]{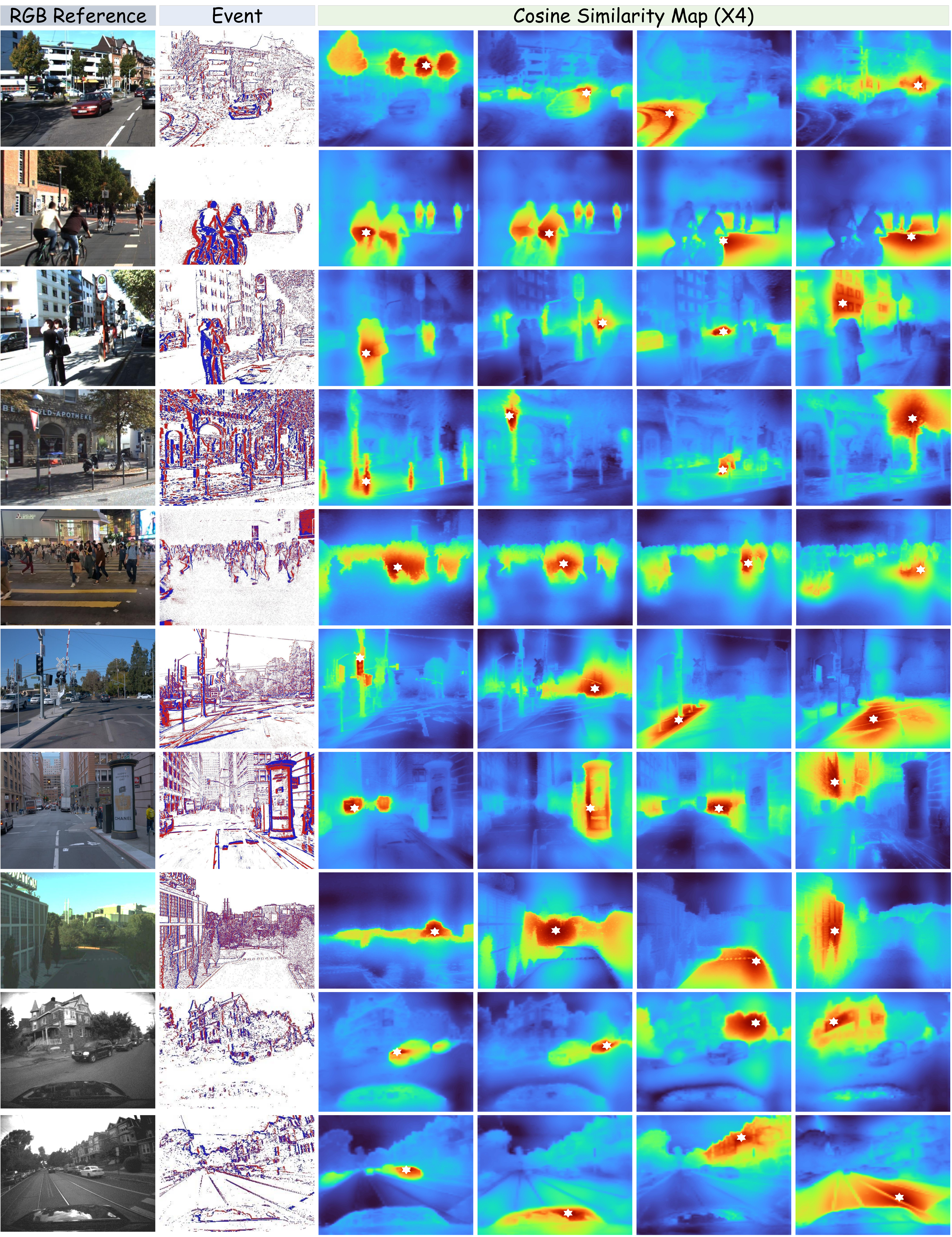}
    \caption{\textbf{The learned fine-grained event features (2/2)} of our method are primarily presented through cosine similarity maps, with key points anchored at the distinct white stars. Best viewed in color.}
    \label{fig:feature-v2}
\end{figure*}

\vskip 0.05in
\noindent \textbf{Exemplary Analysis.} As shown in Figure~\ref{fig:feature-v1} and Figure~\ref{fig:feature-v2}, exemplary learned event features are visualized through cosine similarity maps, with key points marked by white stars. The RGB reference images and corresponding event data are shown on the left, while the cosine similarity maps (scaled by a factor of 4) highlight the areas where the model focuses. These maps emphasize the spatial locations of distinctive event features, demonstrating how the model captures dynamic, fine-grained details. The alignment of the white stars with key features indicates the model's ability to identify significant event-driven changes. The results highlight the model's effectiveness in learning and refining event features, benefiting from the cross-modal distillation of pretrained image-based models to better capture these event features.

\subsection{Downstream Tasks}
\begin{figure*}
    \centering
    \includegraphics[clip, width=0.98\textwidth]{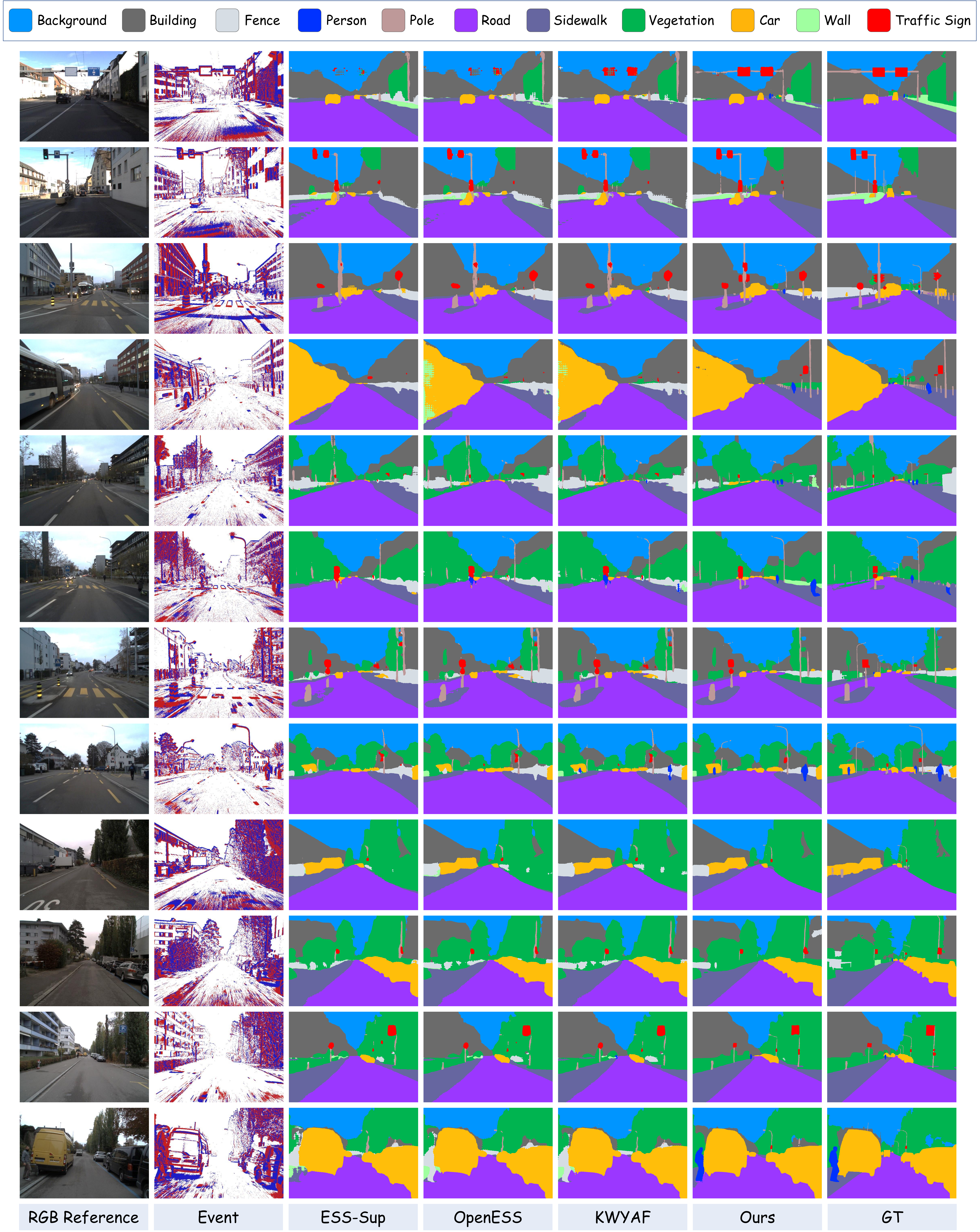}
    \caption{The qualitative comparisons among different \textbf{event-based semantic segmention} approaches on the test set of DSEC-Semantic. Best viewed in color.}
    \label{fig:segment}
\end{figure*}

\begin{figure*}
    \centering
    \includegraphics[clip, width=0.93\textwidth]{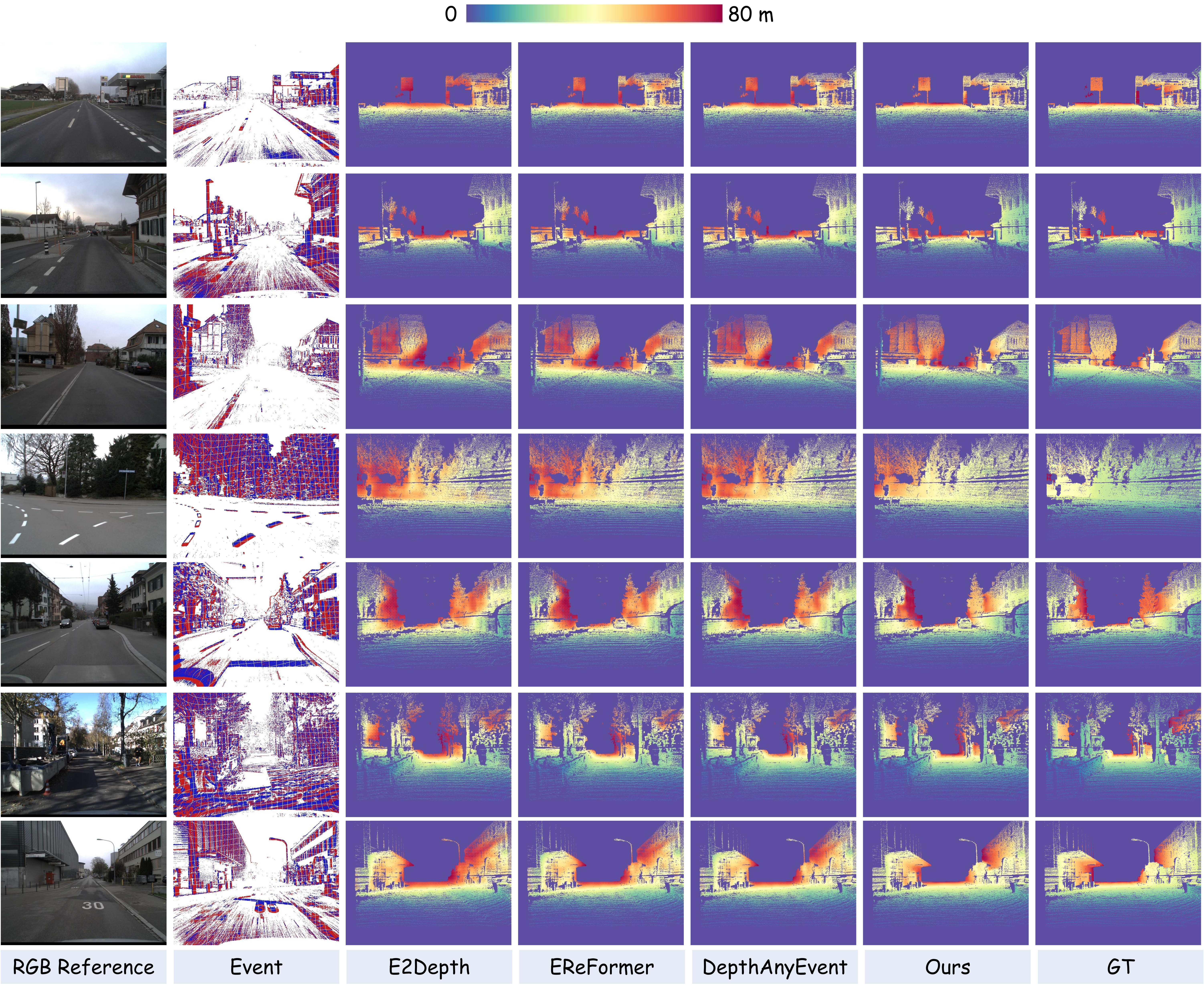}
    \caption{The qualitative comparisons among different \textbf{event-based depth estimation} approaches on the test set of DSEC-Depth. Best viewed in color.}
    \label{fig:depth}
    \vspace{-0.5cm}
\end{figure*}

\begin{figure*}
    \centering
    \includegraphics[clip, width=0.93\textwidth]{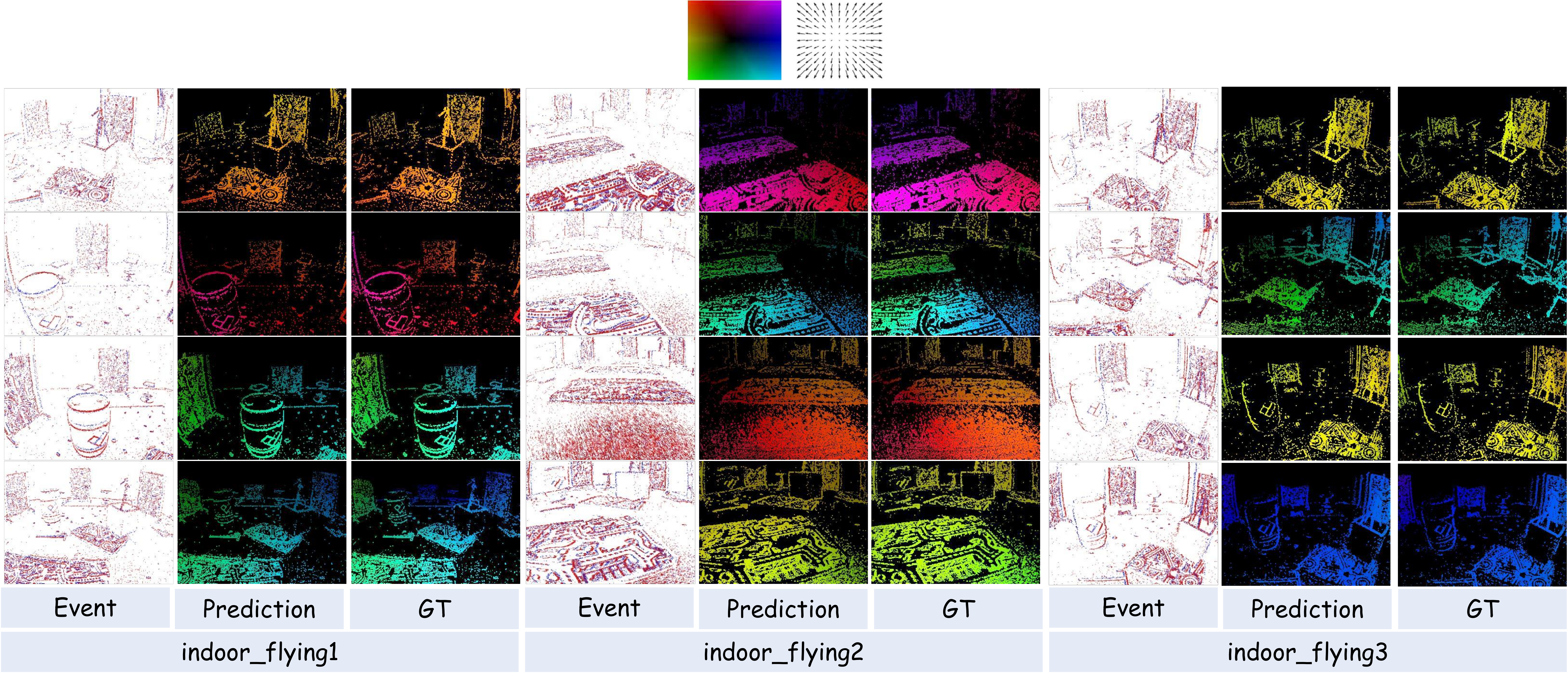}
    \caption{The qualitative results of our \textbf{optical flow estimation} approaches on the test set of MVSEC-Flow. Best viewed in color.}
    \label{fig:flow}
    \vspace{-0.5cm}
\end{figure*}

Representative qualitative results for downstream tasks are provided in Figures~\ref{fig:segment}, ~\ref{fig:depth}, and~\ref{fig:flow}.

\vskip 0.05in
\noindent \textbf{Semantic Segmentation.} As shown in Figure~\ref{fig:segment}, the comparison of event-based semantic segmentation methods on the DSEC-Semantic dataset highlights the effectiveness of cross-modal distillation for dense event pretraining. Our method significantly improves segmentation quality, particularly in fine-grained object boundaries and dynamic features like persons, cars, and traffic signs. The key advantage lies in leveraging pretrained image models through cross-modal distillation, which enhances spatial feature learning in event data. In contrast, methods like ESS-Sup and OpenESS perform well in general segmentation but fail to capture subtle event-driven features, while KWYAF and 6T show some improvement but struggle in dynamic scenes. Our method outperforms them by maintaining high accuracy.

\vskip 0.05in 
\noindent \textbf{Monocular Depth Estimation.} As shown in Figure~\ref{fig:depth}, the comparison of event-based depth estimation methods on the DSEC-Depth dataset demonstrates the benefits of cross-modal distillation for dense event pretraining. Our method produces the most accurate depth maps, especially in dynamic regions with moving objects or occlusions. In contrast, methods like E2Depth and EReformer show noticeable errors, particularly in complex environments. While DepthAnyEvent performs well in static areas, it struggles with depth variations in motion. Our method, leveraging cross-modal pretraining, improves depth accuracy, particularly in foreground-background transitions, by transferring rich spatial knowledge to the event-based depth task.

\vskip 0.05in 
\noindent \textbf{Optical Flow Estimation.} As shown in Figure ~\ref{fig:flow}, the comparison of optical flow estimation results on the MVSEC-Flow dataset highlights the effectiveness of our cross-modal distillation approach. Our method produces the highly accurate and consistent flow predictions, thanks to cross-modal distillation from pretrained models, which enhances flow estimation by leveraging fine-grained correlation knowledge. By transferring knowledge from image-based foundation model, our method improves robustness, capturing fine details and rapid motion changes effectively in event-based data.

\section{Limitation and Discussion}
\label{appendix:imitation}

While our approach significantly advances event-based pretraining, several limitations remain. First, although our structure-aware distillation improves event representation quality, higher resolutions still face some degradation, particularly with patch- and superpixel-level distillation. This suggests that fine-grained alignment methods could be further refined to handle high-resolution event data more effectively. Second, our method relies on large-scale, synchronized image-event datasets, which may not always be feasible to obtain in certain domains. Future work could explore semi-supervised or unsupervised distillation approaches to reduce reliance on these extensive datasets. Additionally, while our model performs well across standard downstream tasks, its ability to generalize to new or rare event-camera configurations remains limited. Addressing this could involve incorporating domain adaptation or meta-learning strategies to improve robustness in more dynamic or occluded environments. Lastly, the computational efficiency of our method, particularly with large encoder models, presents a challenge. Optimizing for lighter backbones or reducing redundant parameters could enhance the applicability of our approach in resource-constrained real-world scenarios, such as robotics or autonomous vehicles.




\clearpage
{
    \small
    \bibliographystyle{ieeenat_fullname}
    \bibliography{main}
}


\end{document}